\documentclass[a4paper,fleqn]{cas-dc}

\usepackage[authoryear]{natbib}
\usepackage{booktabs}
\usepackage{graphicx}
\usepackage{subcaption}
\usepackage{adjustbox}

\def\tsc#1{\csdef{#1}{\textsc{\lowercase{#1}}\xspace}}
\tsc{WGM}
\tsc{QE}
\tsc{EP}
\tsc{PMS}
\tsc{BEC}
\tsc{DE}

\begin{document}
\let\WriteBookmarks\relax
\def\floatpagepagefraction{1}
\def\textpagefraction{.001}

\shorttitle{Hues and Cues}

\shortauthors{Nuria Alabau-Bosque et~al.}

\title [mode = title]{Human-AI Perceptual Alignment by Playing Hues and Cues}                      
\tnotemark[1,2]

\author[1]{Nuria Alabau-Bosque}[type=editor,
                        auid=000,bioid=1,
                        orcid=0009-0002-8092-2066]
\cormark[1]
\fnmark[1]
\ead{nuria.alabau@uv.es}

\credit{Conceptualization, Methodology, Software, Investigation, Writing - Original draft preparation}

\author[1]{Jorge Vila-Tomás}[]
\credit{Data curation, Software, Results analysis, Writing-editing}

\author[1]{Paula Daudén-Oliver}[]
\credit{Data curation, Results analysis, Software, Writing-editing}

\author[1]{Pablo Hernández-Cámara}[]
\credit{Validation, Methodology, Writing-editing}

\author[1]{Valero Laparra}
\credit{Methodology, Validation, Writing-editing, Funding acquisition}

\author[1]{Jesús Malo}

\credit{Data curation, Results analysis, Writing-editing, Funding acquisition}

\affiliation[1]{organization={Image Processing Lab, Universitat de València},
                addressline={Carrer del Catedrátic José Beltrán Martinez}, 
                city={Paterna},
                postcode={46980},
                country={Spain}}

\cortext[cor1]{Corresponding author}

%
\begin{abstract}
Evaluating the perceptual alignment between Contrastive Vision-Language Models (CVLMs) and humans is typically constrained by traditional benchmarks that overlook fine-grained semantic and cultural nuances. In this work, we propose a novel evaluation framework that leverages the gamified, discrete color space of the board game Hues and Cues. By mapping the board's 480 color cells to the CIE xy chromaticity diagram, we calculate empirical perceptual distances across a carefully curated 100-word vocabulary spanning seven semantic categories. To properly contextualize model performance, we establish an empirical lower bound of expected error—the Human Consistency baseline—calculated via Leave-One-Out (LOO) cross-validation on a dense dataset of color associations collected from 325 human observers through a custom digital interface. We evaluate 162 models across multiple architectural families and pre-training datasets to assess their semantic color grounding.  

Our results demonstrate that while CVLMs successfully replicate human cognitive biases, such as idealized memory colors for concrete physical referents (e.g., food and plants), they systematically diverge from the human baseline in abstract, subjective, and pop-culture domains. We identify two distinct failure modes in severely misaligned concepts: semantic misclassification and a systematic uncertainty collapse into a default blue coordinate. Furthermore, we reveal that highly curated pre-training datasets are significantly more effective than massive, uncurated corpora in mitigating these severe misalignments. Ultimately, this work highlights that despite their broad categorization capabilities, current CVLMs still fail to capture the nuanced, localized consensus of human color memory, emphasizing the value of gamified tasks in exposing underlying model biases. The data and code are publicly available to test other metrics.
\end{abstract}


\begin{highlights}
\item A gamified digital interface collected baseline color associations from 325 humans.
\item The 480 discrete board colors were mapped to CIE xy space to evaluate 162 CVLMs.
\item CVLMs systematically collapse to a default blue prior when facing abstract concepts.

\end{highlights}


\maketitle

\section{Introduction}

In recent years, Contrastive Vision-Language Models (CVLMs), pioneered by architectures such as CLIP~\citep{CLIP}, have revolutionized the field of artificial intelligence by demonstrating unprecedented zero-shot capabilities and establishing robust semantic links between visual features and natural language. These models are typically pre-trained on massive, uncurated web-scraped datasets such as LAION~\citep{schuhmann_laion} or YFCC~\citep{thomee_yfcc}. However, despite their impressive performance on broad categorization benchmarks, fundamental gaps remain in their fine-grained perceptual grounding. Recent literature has exposed significant deficiencies in how these models represent and process foundational visual attributes, particularly color, revealing that CVLMs often struggle to form coherent, human-aligned color spaces~\citep{arias_clip}. Furthermore, general studies on model robustness show that when confronted with abstract concepts, CVLMs frequently suffer from semantic hallucinations or distribution collapses~\citep{li_hallucination}, defaulting to majority-class biases rather than acknowledging uncertainty. However, current literature has yet to bridge these two phenomena to explore how semantic ambiguity specifically triggers systematic chromatic collapses within a fine-grained, psychophysically grounded space.

To truly evaluate the human alignment of CVLMs in the chromatic domain, it is essential to look beyond the literal pixel values of objects and consider the cognitive biases that shape human perception. Human color vision is not merely a passive recording of physical wavelengths; it is a highly constructive process. Psychological and psychophysical studies have long established the concept of color diagnosticity~\citep{tanaka_diagnosticity}, demonstrating that object knowledge fundamentally modulates color appearance~\citep{witzel_perception}. This interaction gives rise to the "memory color" phenomenon, where humans typically recall and recognize the prototypical or idealized color of an object rather than its exact physical hue~\citep{hansen_memory, witzel_object}. Furthermore, the way humans categorize and communicate these colors is heavily influenced by semantic and cultural constraints, as evidenced by experimental paradigms like the color communication game~\citep{brown_color_2023}. Consequently, an aligned CVLM should not just predict a physically accurate color, but rather replicate the nuanced, cognitively biased memory color that humans expect.

Traditional computer vision benchmarks, which rely on rigid bounding boxes and static labeling, are ill-equipped to capture these subjective, human-centric nuances. To overcome this limitation, the AI community is increasingly turning towards gamified and open-ended frameworks to evaluate Machine General Intelligence. Playing games is inherently human, requiring reasoning, adaptation, and semantic communication. Recent works have demonstrated that testing Large Language Models and CVLMs within human game environments—whether through repeated game paradigms~\citep{akata_repeteadgames} or scalable game stores designed for cognitive evaluation~\citep{ying_gamestore}—uncovers underlying behavioral patterns, cultural biases, and abstraction failures that traditional testing strategies fail to identify.

In this work, we propose a novel approach to assess the perceptual and semantic color alignment of artificial models via board games. Specifically, we test the color perception capabilities of 162 CVLMs across multiple architectural families and pre-training datasets by simulating the mechanics of the board game Hues and Cues. By mapping the game's discrete 480-cell color board to the standard CIE xy chromaticity diagram, we establish a rigorous framework to measure empirical distances between CVLM predictions and a robust Human Consistency baseline.

Our main contributions are threefold: (1)~We introduce a scalable, gamified evaluation protocol for measuring fine-grained semantic color alignment using standard colorimetric spaces, validated by 325 human observers; (2)~We demonstrate that while CVLMs can replicate human memory colors for concrete physical referents, they suffer a catastrophic divergence in abstract, subjective, and pop-culture domains, frequently resulting in an "uncertainty collapse" towards default chromatic coordinates, akin to previously documented CVLM hallucinations~\citep{li_hallucination}; and (3)~We empirically show that the quality and curation of pre-training datasets play a far more critical role in mitigating these severe perceptual misalignments than sheer architectural complexity. Ultimately, our findings highlight the necessity of deploying human-centric tasks to expose the latent cognitive deficiencies in modern Contrastive Vision-Language Models. All code and data required to reproduce this evaluation are openly available.\footnote{The complete dataset and code repository can be accessed at: \url{https://github.com/Rietta5/HuesAndCues}}"

\section{Methods}
\label{sec:methods}

To bridge the gap between abstract algorithmic evaluation and the human experience, we propose a gamified framework based on the board game Hues and Cues. This environment provides a unique, discrete search space that overcomes the limitations of continuous color pickers, which often hinder direct metric comparisons due to their infinite search space. By mapping the game's standardized $16\times30$ color grid to the CIE$xy$~\citep{wyszecki_color} chromaticity diagram, we transform subjective board coordinates into an objective, colorimetric space. This section outlines our experimental design, detailing the human data acquisition process, the statistical metrics employed to quantify perceptual alignment—including Mahalanobis distance~\citep{Mahalanobis36} for outlier detection and Hotelling’s $T^2$~\citep{Hotelling} test for distributional assessment—and our protocol for detecting semantic uncertainty in Contrastive Vision-Language Models.

\subsection{Hues and Cues Environment}
\label{sec:huesEnviroment}

To evaluate the alignment between human perception and AI models in a controlled setting, we utilize the color space defined by the board game Hues and Cues. Unlike continuous color pickers (e.g., a full RGB spectrum), which present an infinite search space and hinder direct comparison, this environment provides a discrete and standardized search space.

\begin{figure*}[!h]
	\centering
	\includegraphics[width=1\linewidth]{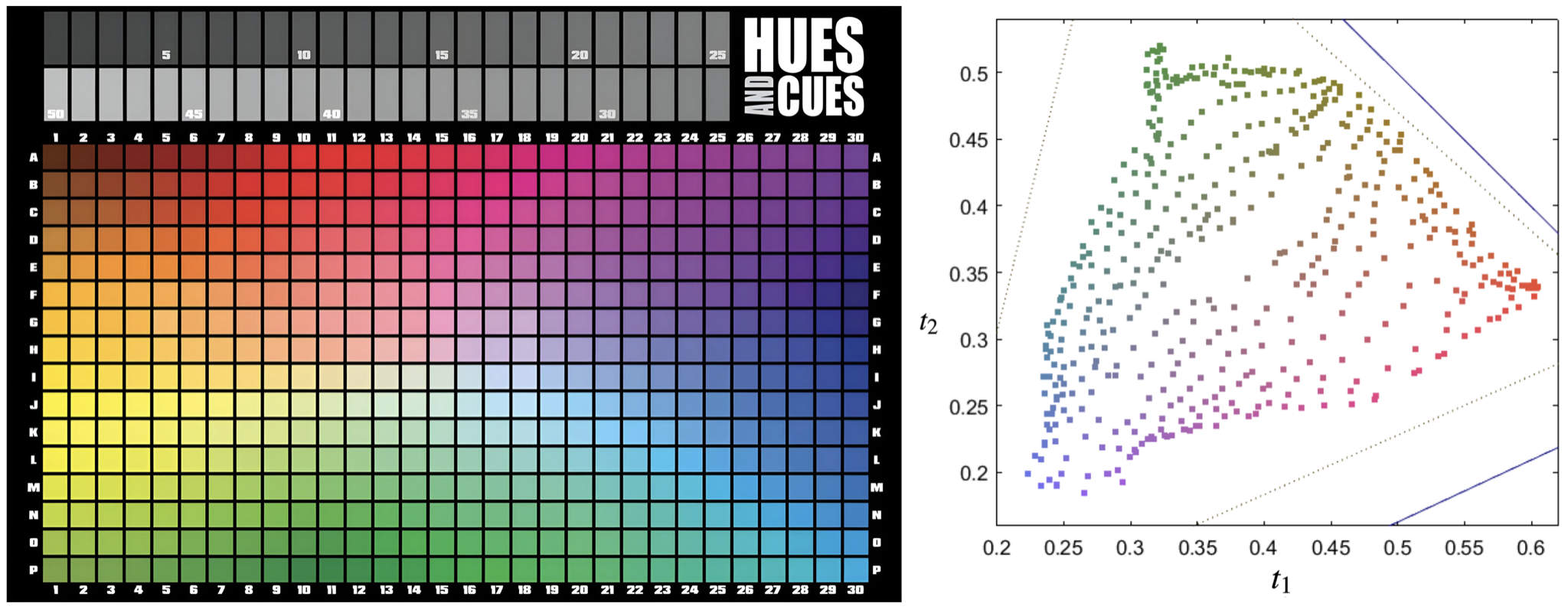}
	\caption{Definition of the discrete color space used in the study. Left: The Hues and Cues game board, consisting of a $16 \times 30$ grid of unique colors (Total $N=480$). Right: Projection of the 480 board colors onto the CIE 1931 $xy$ chromaticity diagram. Note how the regular grid structure becomes distorted when mapped to the perceptual space.}
	\label{fig:Colores}
\end{figure*}

The "board" acts as a quantization matrix composed of $N=480$ unique colors, organized in a grid of 16 rows (labeled A to P) by 30 columns (numbered 1 to 30), as shown in the left panel of Figure~\ref{fig:Colores}. The spatial arrangement of the grid is not arbitrary but approximates the topological distribution of the CIE$xy$ chromaticity diagram. Specifically, the vertical axis roughly corresponds to the Red-Green opponent channel, while the horizontal axis aligns with the Blue-Yellow gradient, effectively creating a discretized map of the color spectrum.

To perform rigorous metric comparisons that transcend the arbitrary grid coordinates (e.g., "C12"), it is necessary to translate this discrete space into a measurable color space. We mapped each of the 480 board cells to their corresponding coordinates in the CIE$xy$ chromaticity diagram.

The right panel of Figure~\ref{fig:Colores} illustrates this transformation. Each point represents one of the 480 board colors projected onto the 2D chromaticity space. Note that the grid, which appears regular on the physical board, it is a non-uniform and curvilinear distribution in the $xy$ space. 


In both human data collection (Section~\ref{sec:humanAcquisition}) and model evaluation (Section~\ref{sec:evaluation}), the task is restricted to selecting one (or a set) of these 480 discrete points as the best visual representation of a given semantic concept.

\subsection{Human Baseline Acquisition}
\label{sec:humanAcquisition}

To establish a dense probabilistic "ground truth," we developed a cross-platform web~\footnote{The gamified data collection interface is publicly accessible at: \url{https://huesandcues-d76ee.web.app}} interface optimized for both desktop and mobile devices. Crucially, the digital board displayed to users was reconstructed cell-by-cell using the colorimetric measurements detailed in Section~\ref{sec:ModelBenchmarking}, ensuring that the digital source of the stimuli presented to humans matches the inputs processed by the models. Our data collection protocol prioritizes ecological validity and large-scale acquisition over the strictly controlled conditions of a laboratory setting, resulting in a final cohort of 325 human participants.

Unlike traditional psychophysical studies where viewing conditions (lighting, monitor calibration, distance) are standardized, our participants performed the task on their personal devices. This introduces variance in color rendering (gamma, gamut coverage). However, we assume that similarly to color constancy under change of illumination~\cite{fairchild_color}, human chromatic adaptation  
naturally compensates for these shifts, so semantic color association is invariant to these changes. To empirically validate this premise, Appendix~\ref{app:Adaptation} demonstrates that color variance across uncalibrated displays is statistically equivalent to natural illuminant changes, which do not disrupt color categorization. For instance, while a specific shade of "Red" for the word \texttt{CARNATION} may shift in chromatic coordinates due to uncalibrated screens, this shift is compensated so color category is inveriant (e.g., Green or Blue). This trade-off allows us to capture data from a naturally distributed population rather than a small, localized sample.

\begin{figure*}[!h]
	\centering
	\includegraphics[width=1\linewidth]{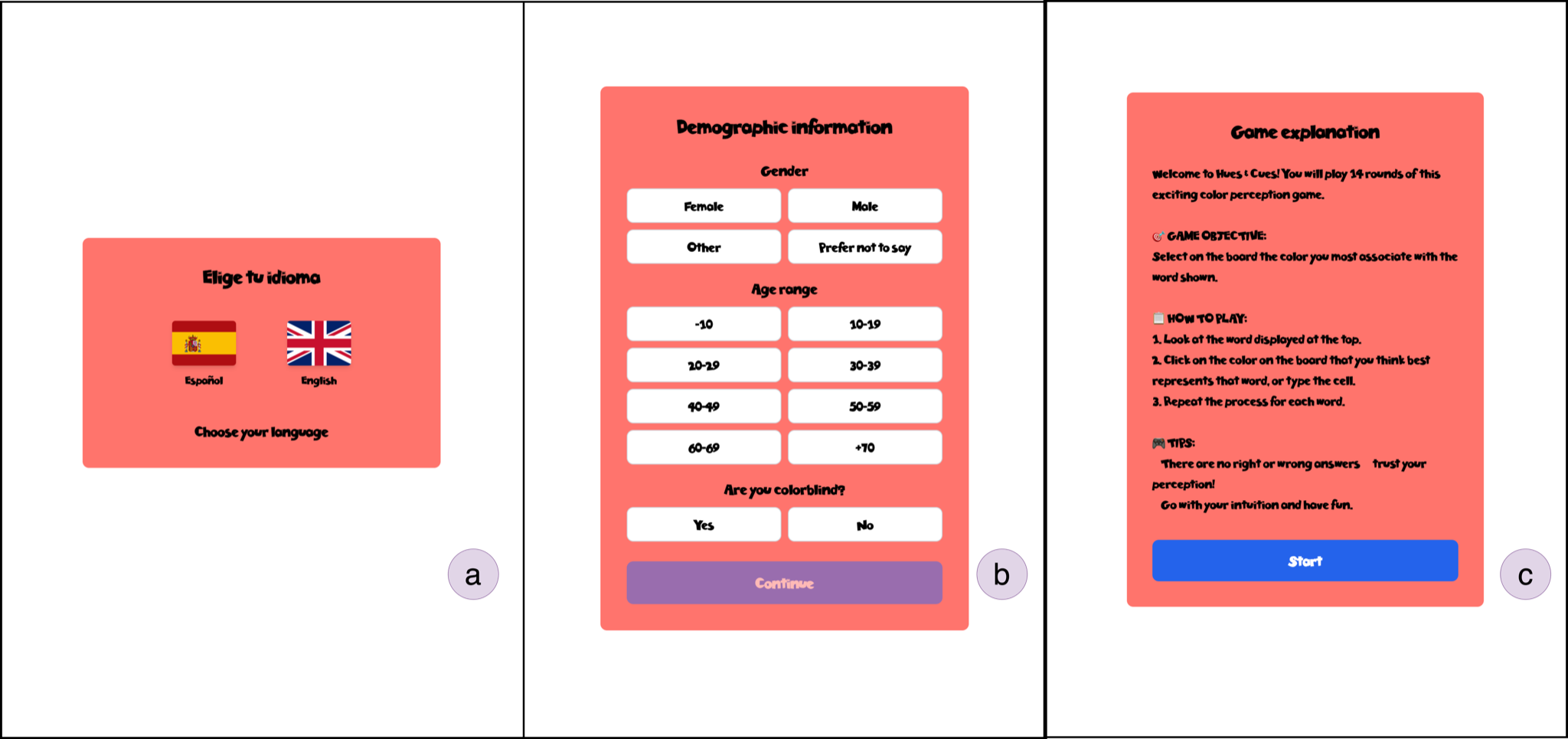}
	\caption{User onboarding and demographic profiling. (a)~Language selection allows for cross-cultural analysis of specific terms. (b)~Self-reported screening for gender, age and Color Vision Deficiency (CVD). (c)~Instructions emphasizing intuitive perception.}
	\label{fig:Datos}
\end{figure*}

Since direct optometric validation (e.g., visual acuity, color blindness tests) is not feasible in a remote setting, we rely on self-reported metadata. As shown in Figure~\ref{fig:Datos}, the onboarding process includes a specific screening for Color Vision Deficiency (CVD). This allows us to filter or segment responses from users with anomalous trichromacy or dichromacy during the analysis phase. Additionally, demographic data (age, gender, language) is collected to enable cross-cultural comparisons.

\begin{figure*}[!h]
	\centering
	\includegraphics[width=1\linewidth]{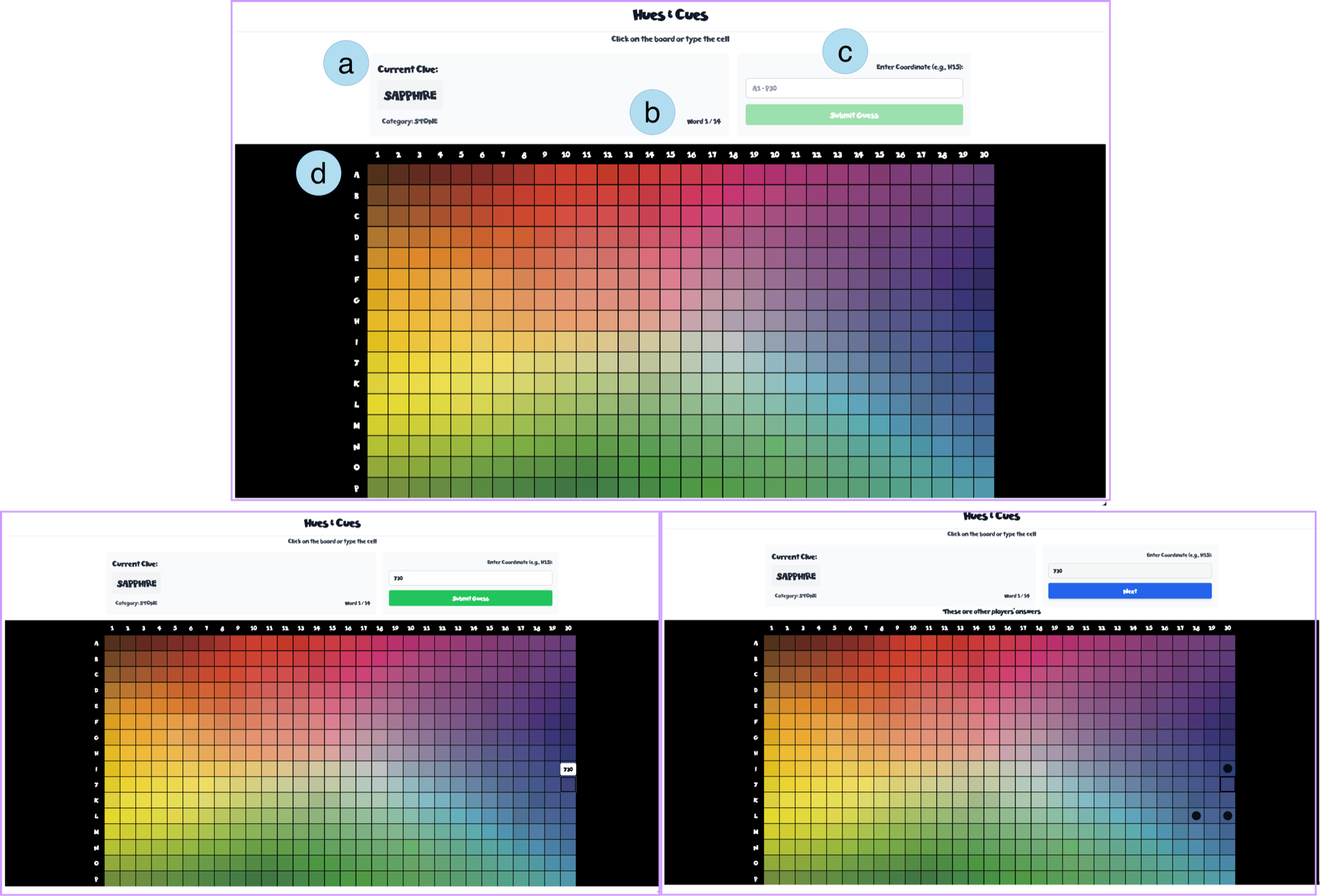}
	\caption{Data collection interface details. Blue markers indicate specific UI components: (a)~Target stimulus presentation (Current Cue). (b)~Trial progress indicator. (c)~Text input field for coordinate entry (alternative to clicking). (d)~The interactive $16 \times 30$ color grid. The panels on the right illustrate the submission step and the Social Feedback phase, where the distribution of previous answers is revealed (black markers) after the user's choice.}
	\label{fig:Web}
\end{figure*}

We curated a dataset of 100 words stratified into seven semantic categories: Animal, Food, Mineral, Plant, Environment, Subjective, and Pop Culture. The selection criteria prioritized two objectives:(1)~Words were chosen to span the entire board, preventing color imbalances. (2)~We introduced "cultural anchors" to test if associations are driven by visual reality or cultural memes. For example, \texttt{TRACTOR} was included for Spanish speakers (referencing a popular song about a "yellow tractor," despite real tractors often being red/green), and \texttt{SUBMARINE} for English speakers ("Yellow Submarine").
To minimize cognitive fatigue, the 100-word vocabulary is randomized, presenting each user with a sequential subset of 15 words. The interaction flow (Figure~\ref{fig:Web}) follows a forced-choice protocol:

\begin{itemize}
    \item The target word is displayed~(a), along with the current progress~(b). 
    \item The user maps the concept to a specific coordinate on the grid~(d), either by clicking directly or typing the coordinate~(c).
    \item  Post-submission, the interface reveals aggregated responses from previous users (shown in the bottom-right panel of Figure~\ref{fig:Web}), incentivizing participation without biasing the initial choice.
\end{itemize}

\subsection{Human Dataset Characteristics and Validation}
\label{sec:human_bbdd}

\begin{figure*}[!h]
	\centering
	\includegraphics[width=1\linewidth]{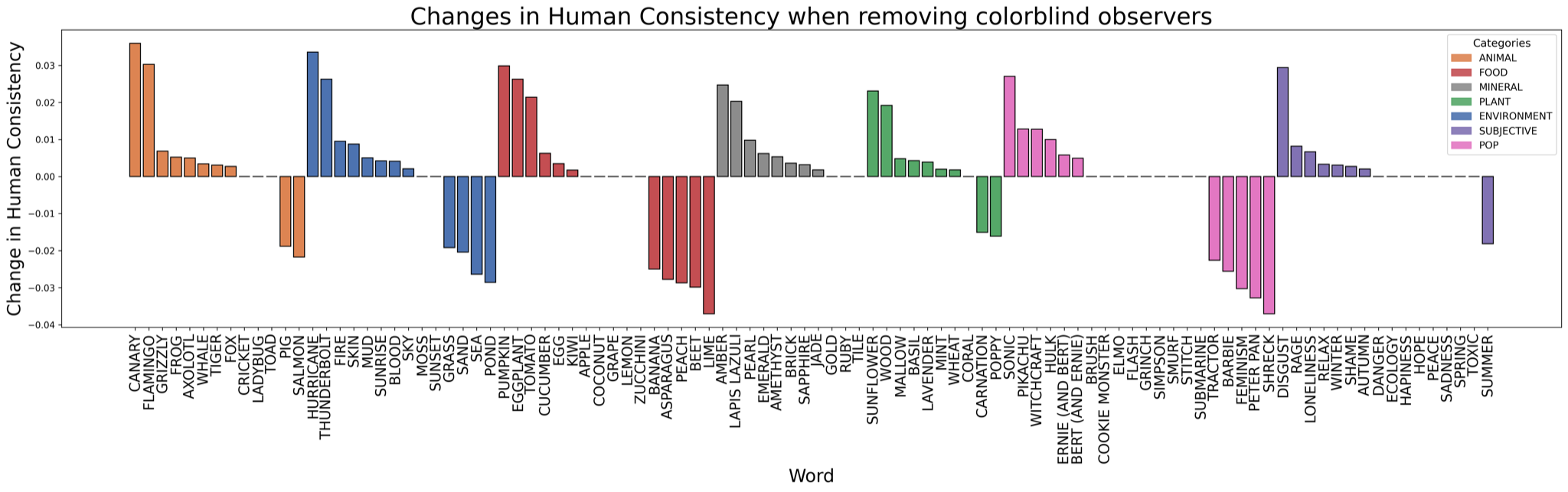}
	\caption{Impact of Color Vision Deficiency (CVD) on dataset consistency. The bar chart illustrates the absolute change in the Human Consistency metric for each word after excluding the 9 self-reported colorblind participants. While the magnitude of change is generally small across most categories, their removal effectively isolates normal trichromatic consensus and prevents anomalous outliers in semantic regions typically affected by color blindness.}
	\label{fig:Daltonicos}
\end{figure*}

To establish a robust empirical baseline, we analyzed the responses collected from our crowdsourced digital interface, which initially comprised baseline color associations from 325 human observers. To guarantee the integrity of the perceptual ground truth, we applied a strict filtering criterion based on self-reported visual capabilities, excluding 9 participants who reported Color Vision Deficiency (CVD) during the onboarding phase. To quantify the impact of these anomalous color associations on our dataset, we evaluated the change in the Human Consistency metric before and after their exclusion. As depicted in Figure~\ref{fig:Daltonicos}, while the overall deviation introduced by CVD participants is relatively minor, their removal systematically refines the probabilistic consensus across the vocabulary. To provide further transparency, we have developed an interactive web visualization~\footnote{Readers can visually explore and compare the specific color associations made by CVD participants versus normal trichromats at: \url{http://147.156.207.27:5000/ }} that allows readers to independently explore the specific board coordinates selected by colorblind observers for each concept. Ultimately, this refinement resulted in a final validated cohort of 316 normal trichromat observers.

\begin{figure*}[!h]
	\centering
	\includegraphics[width=1\linewidth]{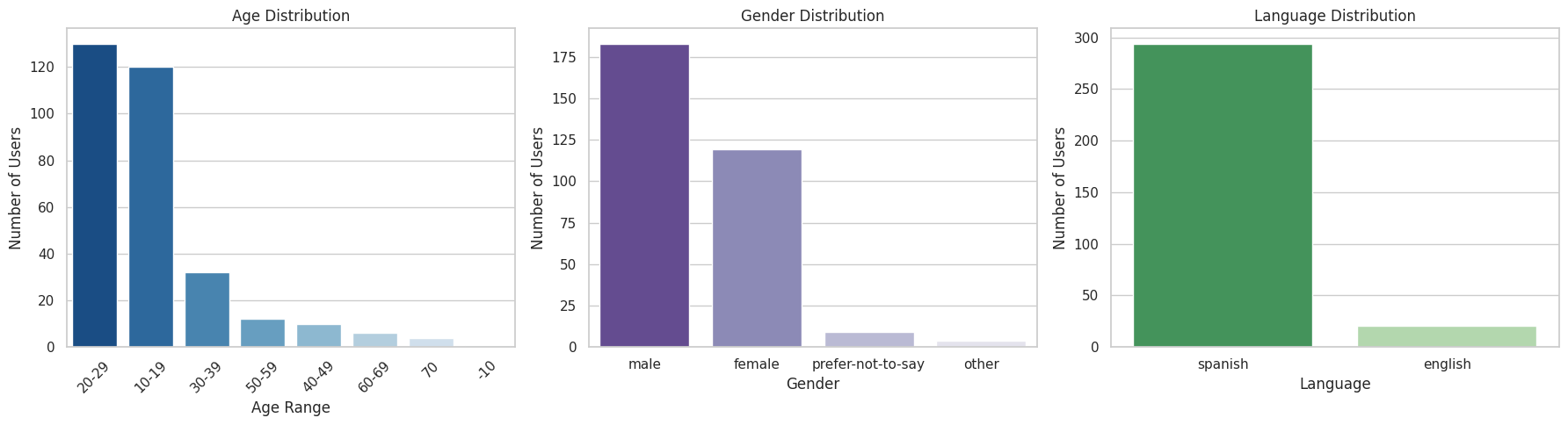}
	\caption{Demographic profile of the validated human cohort. The distributions detail participant age, gender, and language, highlighting a predominantly young, Spanish-speaking participant base.}
	\label{fig:Demografico}
\end{figure*}

\begin{figure*}[!h]
	\centering
	\includegraphics[width=1\linewidth]{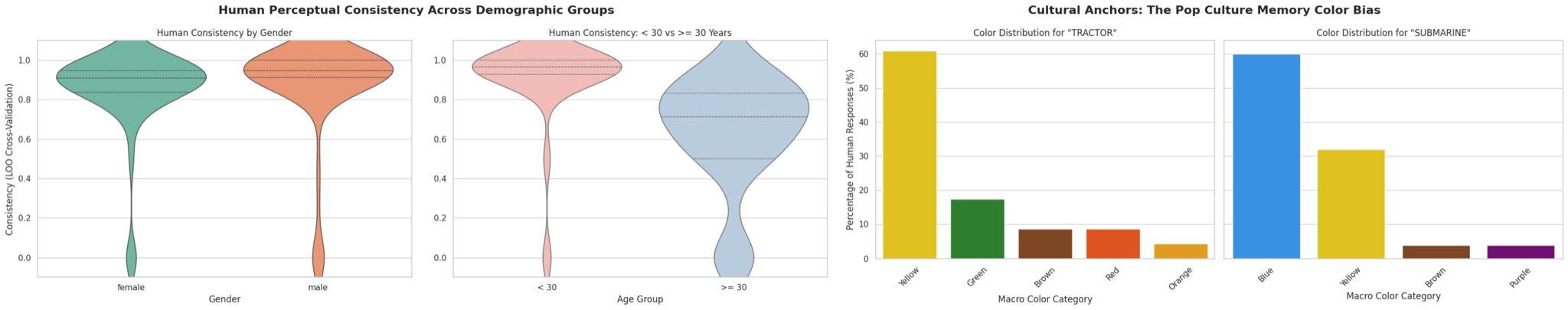}
	\caption{Left: Violin plots demonstrate that perceptual consistency remains robust and statistically stable regardless of gender or age variations. Right: Color category distributions for the cultural anchors \texttt{TRACTOR} and \texttt{SUBMARINE}SUBMARINE, highlighting the strong influence of localized pop culture on semantic color associations.}
	\label{fig:DemogCons}
\end{figure*}

Figure~\ref{fig:Demografico} details the demographic composition of our final cohort. The participant pool presents a predominantly young demographic, with the majority of users falling into the 10-19 and 20-29 age brackets. Gender distribution reflects a higher proportion of male respondents compared to females, and the language breakdown is heavily weighted towards Spanish speakers, alongside a smaller English-speaking subset. Crucially, as demonstrated in the left panels of Figure~\ref{fig:DemogCons}, these demographic imbalances do not introduce significant perceptual bias into our ground truth. Human consistency scores remain remarkably stable and statistically indistinguishable across different gender and age boundaries (<30 vs. $\geq$30).  

Beyond concrete physical referents, our vocabulary included specific "cultural anchors" designed to test the influence of localized pop culture on color memory. The right panels of Figure~\ref{fig:DemogCons} illustrate this phenomenon. For instance, despite real-world agricultural tractors typically being green or red, the Spanish-speaking majority overwhelmingly associated the word \texttt{TRACTOR} with the color yellow. Similarly, the term SUBMARINE exhibits a notable split between literal physical associations (blue) and cultural ones (yellow). However, the influence of the "Yellow Submarine" pop-culture reference is currently weaker in our dataset. As indicated by Figure~\ref{fig:Demografico}, this dilution directly correlates with the low representation of English speakers in the current cohort. Because the gamified interface remains continuously active for data collection, future efforts will focus on viralizing the platform among currently underrepresented segments—such as diverse language speakers and older age groups—to progressively balance the demographics and capture a more universal consensus.

\begin{figure*}[!h]
	\centering
	\includegraphics[width=1\linewidth]{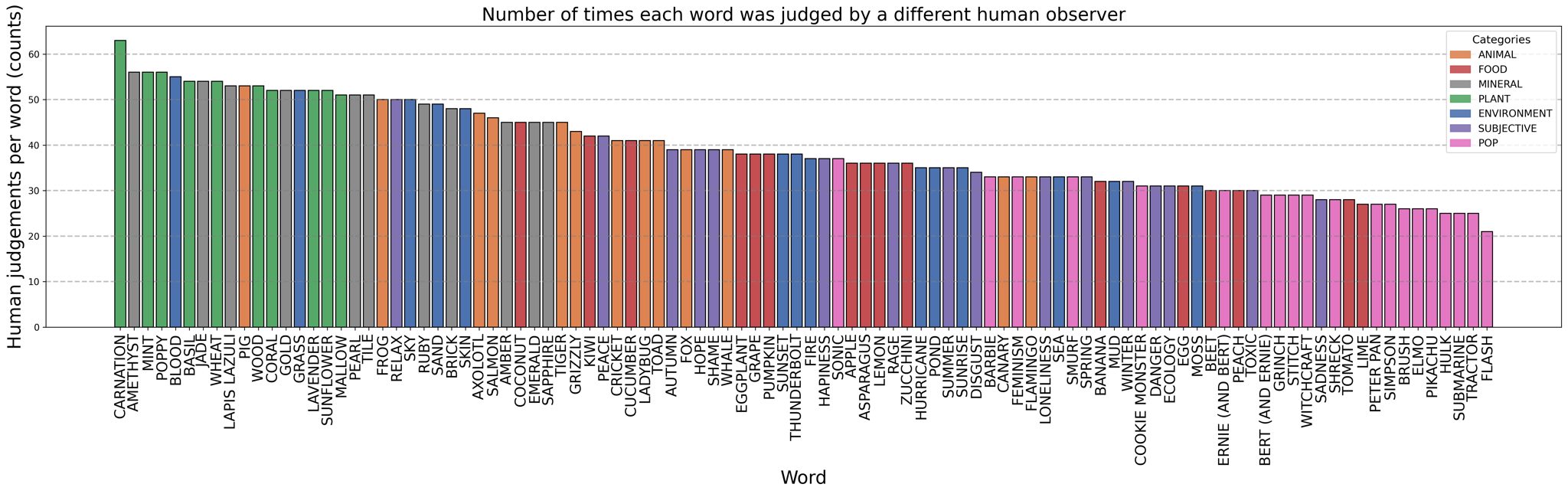}
	\caption{Distribution of human responses across the curated 100-word vocabulary. The bar chart illustrates the total number of unique human judgments collected per word, ordered by descending frequency. Colors denote the predefined semantic category of each concept.}
	\label{fig:ConteoPalabras}
\end{figure*}

As detailed in Section~\ref{sec:humanAcquisition}, to prevent cognitive fatigue, each participant was presented with a randomized subset of 15 words from the total 100-word vocabulary. Figure~\ref{fig:ConteoPalabras} illustrates the resulting distribution of human judgments per concept. Despite the randomized assignment, this sampling strategy ensured a dense and statistically viable number of annotations across the entire evaluation set. Every concept received a minimum of 20 independent associations, with the most frequent terms exceeding 50 judgments, thereby providing a solid, high-density foundation for calculating the Human Consistency baseline.

\subsection{Model Benchmarking Setup}
\label{sec:ModelBenchmarking}

To assess the alignment capability of current Contrastive Vision-Language Models (CVLMs) across different architectures and training regimes, we scaled the experiment to a comprehensive suite of 162 CLIP-based models. This selection includes variants trained on diverse datasets (e.g., WebLI, LAION-400M, LAION-2B, DataComp) and utilizing various backbones (e.g., ResNet, ViT-B/32, ViT-L/14, ConvNeXt). To facilitate a structured analysis of this vast evaluation space, we categorized the 162 checkpoints into distinct architectural and dataset families. Architecturally, models were grouped based on their core design principles and contrastive loss mechanisms (e.g., classical OpenAI architectures, the SigLIP family, EVA, and MobileCLIP). From a data-centric perspective, we classified the models according to the curation philosophy of their pre-training corpora, distinguishing between massive, uncurated web-scraped pools (e.g., the LAION family) and highly filtered, curated datasets (e.g., DataComp, DFN, and Meta's CommonPool). Table~\ref{tab:tabla_conteo} summarizes the quantitative distribution of the evaluated models across these architectural and dataset categories. A comprehensive table mapping each of the 162 evaluated models to its corresponding architectural and dataset family is provided in Appendix~\ref{app:ListadoCompleto}.

\begin{table}[h]
\centering
\caption{Distribution of the evaluated models grouped by architecture family and pre-training dataset origin.}
\label{tab:tabla_conteo}
\begin{tabular}{@{} l r r @{}}
\toprule
\textbf{Grouping by model} & \textbf{Count} & \textbf{Percentage (\%)} \\
\midrule
OpenAI                     & 94 & 58.0 \\
SigLIP                     & 30 & 18.5 \\
CoCa / ConvNeXt            & 16 &  9.9 \\
MobileCLIP                 & 10 &  6.2 \\
EVA                        & 7  &  4.3 \\
Multilingual               & 5  &  3.1 \\
\midrule
\textbf{Total}             & \textbf{162}& \textbf{100.0} \\
\midrule
\textbf{Grouping by dataset} & \textbf{Count} & \textbf{Percentage (\%)} \\
\midrule
Owners (WebLI)              & 49 & 30.2 \\
LAION (Massive scale)       & 39 & 24.1 \\
DataComp / DFN (Curated)    & 37 & 22.8 \\
Meta (CommonPool/MetaCLIP)  & 34 & 21.0 \\
Pioneers (YFCC/CC)          & 3  &  1.9 \\
\midrule
\textbf{Total}             & \textbf{162}& \textbf{100.0} \\
\bottomrule
\end{tabular}
\end{table}

To eliminate discrepancies between the human and AI search spaces, we standardized the color stimuli used throughout the experiment. To obtain the reference colors, we displayed the official board image provided by the publisher and measured the chromatic coordinates (CIE $xy$) of each cell using a calibrated colorimeter. This empirical measurement was necessary because inherent pixel variance in the source image prevented the direct extraction of a single, homogeneous color value per patch. These measurements were subsequently transformed into the standard sRGB space. By utilizing these standardized digital values to generate both the discrete image patches processed by the models and the "Digital Twin" of the board for the human web interface, we ensure that the evaluation framework originates from the exact same colorimetric source for both biological and artificial agents.

We define the retrieval task as finding the visual representation that maximizes the semantic similarity with the text prompt. For each word $w$ in our 100-word vocabulary, the process is as follows: First, the word $w$ is tokenized and passed through the text encoder of the model $\mathcal{M}$ to obtain a text embedding $T_w$. Simultaneously, the 480 measured color patches are processed by the image encoder to obtain a set of visual embeddings $\{I_1, I_2, ..., I_{480}\}$. Second, we compute the cosine similarity between the text embedding and every visual embedding in the search space. And finally, the colors are ranked based on their similarity scores. We collect the Top-5 ($k=5$) candidates for each word/model pair. This allows us to analyze the model's "confidence cloud" and compare it with the distribution of human responses.

\subsection{Evaluation Metrics}
\label{sec:evaluation}

To quantify the alignment between human perception and CLIP models, we employ a two-tiered statistical approach, analyzing both the models' "best guess" (Top-1) and their "uncertainty cloud" (Top-5).

\subsubsection{Point-wise Validity (Outlier Analysis)}
\label{sec: outlier}

First, we assess whether the model's primary prediction (Top-1) falls within the realm of human consensus. Since human color associations typically form anisotropic clusters (ellipses) rather than perfect circles in the CIE $xy$ space, Euclidean distance is an insufficient metric. Instead, we utilize the Mahalanobis Distance ($D_M$)~\citep{Mahalanobis36}. For each word, we model the human responses as a bivariate normal distribution defined by a mean vector $\boldsymbol{\mu}$ and a covariance matrix $\mathbf{\Sigma}$. The distance of a model's prediction $\mathbf{x}$ is calculated as in the Equation~\ref{ec:Mahalanobis}.

\begin{equation}
    D_M(\mathbf{x}) = \sqrt{(\mathbf{x} - \boldsymbol{\mu})^T \mathbf{\Sigma}^{-1} (\mathbf{x} - \boldsymbol{\mu})}
    \label{ec:Mahalanobis}
\end{equation}

A model's response is classified as an outlier if its $D_M$ exceeds the critical value derived from the Chi-square distribution ($\chi^2_2$) at a 95\% confidence level. This test filters out the cases where the model's prediction is statistically incompatible with the human distribution.

\subsubsection{Distributional Alignment (Hotelling’s Two-Sample Test)}
\label{sec: hotteling}

While the Mahalanobis metric evaluates individual outliers, we also assess whether the collective response of the model aligns with the human consensus. We treat the model's Top-5 responses as a sample vector set and compare it against the human response set using Hotelling’s Two-Sample $T^2$ test~\citep{Hotelling}. This test serves as the multivariate generalization of the Student's t-test, operating directly in the 2D chromatic diagram. The null hypothesis ($H_0$) posits that the mean centroid of the model's predictions ($\boldsymbol{\mu}_M$) is statistically indistinguishable from the human centroid ($\boldsymbol{\mu}_H$). We used a significance level of $p = 0.05$ to reject $H_0$. 

\subsubsection{Human Consistency Baseline}
\label{sec: Consistencia_humana}

To properly contextualize the performance of the evaluated models, it is essential to establish an empirical baseline of human perception. Human color perception and semantic association are inherently noisy processes; therefore, assuming a perfect, zero-error ground truth is unrealistic.  

To quantify this internal agreement, we compute a "Human Consistency" metric using a Leave-One-Out (LOO) validation approach. For each target word $w$ with a total of $N$ human responses, we isolate a single human response $x_i$. We then estimate the distribution parameters (mean vector $\mu$ and covariance matrix $\Sigma$) using the remaining $N-1$ responses. The isolated response $x_i$ is subsequently evaluated against this $N-1$ distribution using the exact same statistical criteria applied to the models (e.g., the Mahalanobis distance outlier detection described in Section~\ref{sec: outlier}).  

This cross-validation process is repeated iteratively for all $N$ responses across all words in our vocabulary to determine the overall Human Consistency. However, to facilitate direct comparison with the models, whose results are often reported in terms of error, we translate this metric into a "Human Fail Rate" (or human error rate), calculated as:  $\text{Human Fail Rate} = 1 - \text{Human Consistency}$

This failure rate represents the inherent noise or semantic disagreement within the human population itself. It serves as the empirical lower bound of expected error. Consequently, a model is considered to achieve optimal perceptual alignment not when its error rate reaches absolute zero, but when it converges towards this baseline of human intra-class variability.

\subsection{Uncertainty Protocol and Bias Detection}
\label{sec: colapso}

A critical aspect of evaluating Contrastive Vision-Language Models is distinguishing between misalignment (associating a concept with the wrong color) and uncertainty (having no strong association and reverting to a prior). We hypothesize that when CLIP models face semantic ambiguity or out-of-distribution concepts, they do not sample uniformly from the color space but collapse towards a specific chromatic region driven by training data imbalances.

To isolate this behavior, we designed a "Nonsense Baseline" test. In addition to the 100-word vocabulary, we subjected the 162 models to a set of 16 semantically void inputs. To systematically evaluate different types of semantic absence, these inputs were evenly distributed into four distinct categories (four realizations per category), directly mirroring the matrix structure evaluated in Figure 8:
\begin{itemize}
    \item Numbers: Abstract numerical digits with no inherent chromatic association.
    \item Stop Words: High-frequency functional language tokens (e.g., "the", "and", "of", "is").
    \item Pseudowords: Pronounceable but structurally meaningless terms (e.g., "Glimborp").
    \item Random Strings: Pure alphanumeric noise (e.g., \newline"xfhjk\_1").

\end{itemize}

Previous studies on large-scale image datasets~\citep{schuhmann_laion} (e.g., LAION, CommonCrawl) suggest a prevalence of "cold" tones due to the overrepresentation of outdoor scenes (sky, water) and corporate web design. We posit that under conditions of maximum uncertainty (nonsense prompts), models will exhibit a "Blue Bias," systematically selecting coordinates in the lower-left quadrant of the CIE $xy$ diagram (blue-violet region) rather than exhibiting high-entropy (random) behavior. This protocol allows us to quantify the model's intrinsic prior. If a model's error pattern for valid words (e.g., \texttt{HOPE}) aligns with its response to nonsense inputs, the error can be attributed to uncertainty collapse rather than a learned semantic misassociation.

\subsection{Color Categorization Mapping}
\label{sec: color_naming}

While the Hues and Cues board provides a precise $16 \times 30$ coordinate system for empirical measurement, these alphanumeric labels (e.g., A13, P10) lack direct interpretability for qualitative analysis. To bridge the gap between continuous chromatic coordinates and natural language, we established a mapping between the discrete board space and basic human color categories.  

\begin{figure*}[!h]
	\centering
	\includegraphics[width=1\linewidth]{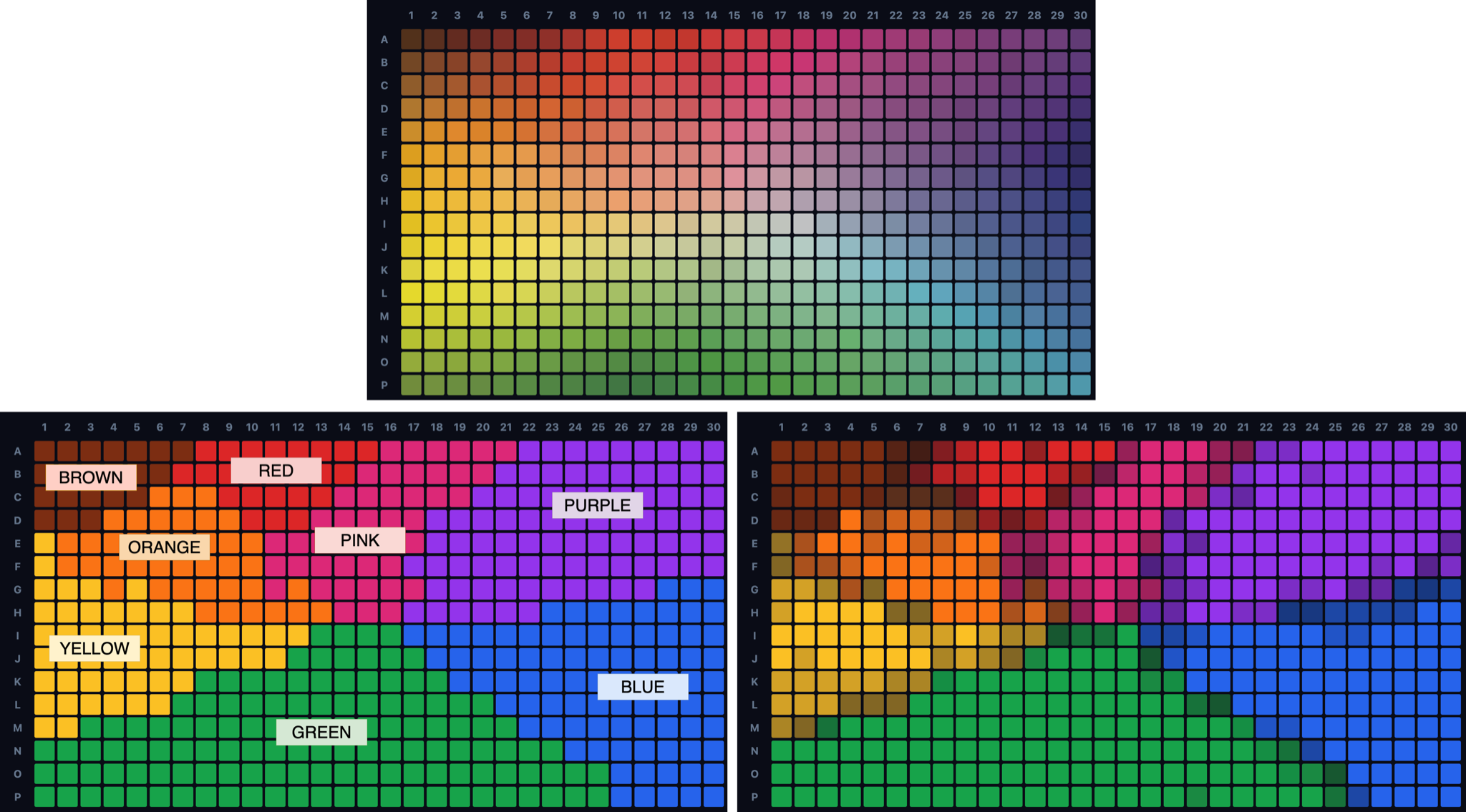}
	\caption{Tessellation of the Hues and Cues playing board into eight fundamental English color categories. Top: The original continuous discrete color grid. Bottom-left: The final categorical mapping translating raw coordinates into interpretable color names for qualitative comparison. Bottom-right: The human consensus map, where darker or blended cells highlight the decision boundaries, reflecting the natural perceptual uncertainty among observers when defining the edges of basic color regions.}
	\label{fig:ColorNaming}
\end{figure*}

As illustrated in Figure~\ref{fig:ColorNaming}, we tessellated the 480-cell board into eight fundamental color regions: Brown, Red, Orange, Yellow, Green, Blue, Purple, and Pink. To establish the definitive boundaries of these regions, we developed an interactive web application~\footnote{Readers can explore and participate in the interactive color categorization mapping at: \url{http://147.156.207.27:8000}} that allows users to manually divide the board into these basic color areas. For the initial regionalization used in this study, we derived the consensus from a subset of eight human participants. The final regionalization was derived from the consensus of their divisions, which naturally aligned with the focal colors commonly established in color naming literature (e.g., the Berlin and Kay basic color terms \cite{berlin_kay_1969}). The bottom-right panel of Figure~\ref{fig:ColorNaming} visualizes this resulting consensus, explicitly exposing the decision boundaries; cells with lower agreement appear darker or blended, successfully capturing the inherent perceptual ambiguity between adjacent color categories.

This macro-categorization enables a secondary level of evaluation. Beyond calculating the mathematical distance to the human centroid (as per our Mahalanobis metric), we can assess the semantic severity of a model's error. For instance, predicting a slightly lighter shade of green for \texttt{CUCUMBER} is a minor coordinate error, whereas predicting a blue coordinate constitutes a severe categorical failure (semantic bleeding). By translating both the human consensus coordinates and the models' predictions into these eight textual categories, we facilitate a direct, word-level qualitative comparison.

\section{Results}
\label{sec:results}

We assess the perceptual and semantic alignment of 162 Contrastive Vision-Language Models by evaluating their color-naming capabilities across a curated 100-word vocabulary. The analysis presented in this section evaluates model performance through two lenses: global perceptual consistency across architectural families and datasets, and fine-grained categorical analysis. To benchmark these findings, we utilize the 'Human Consistency' score as an empirical lower bound for expected error. We demonstrate that while model performance is highly influenced by pre-training corpus curation, a pervasive 'uncertainty collapse' persists, highlighting a systemic divergence between the models' confidence clouds' and the localized consensus of human memory color.

\subsection{Global Model Performance}
\label{sec:results_global}

\begin{figure*}[t]
    \centering
    \includegraphics[width=1\linewidth]{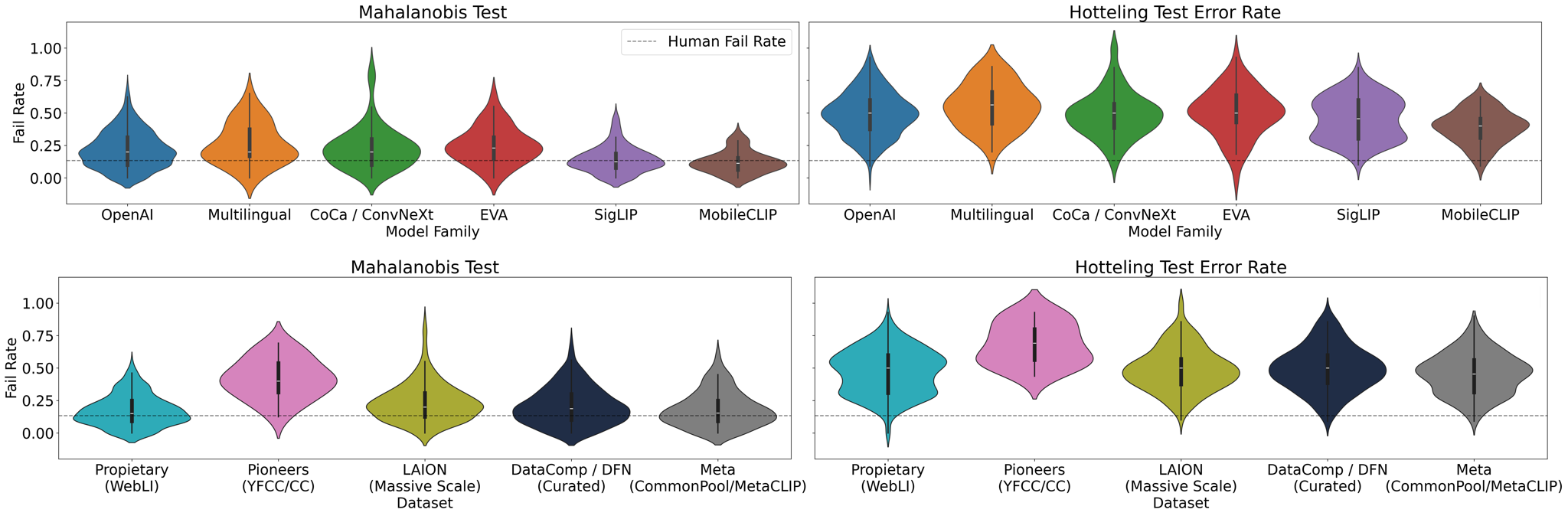}
    
   \caption{Global performance and perceptual alignment of the 162 evaluated models with respect to the human memory color baseline. The top row illustrates the error distributions grouped by architectural family, while the bottom row presents the same distributions grouped by pre-training dataset origin. Left column: Mahalanobis distance from the human consensus. Right column: Hotelling's $T^2$ test error rates. In all panels, the horizontal dashed line represents the human baseline performance.}
    \label{fig:cuadricula_violines_completa}
\end{figure*}

We first evaluate the global perceptual alignment of the 162 models across the entire vocabulary. Figure~\ref{fig:cuadricula_violines_completa} summarizes these results, analyzing the error distributions both by architectural family and by pre-training dataset origin.

To properly interpret these distributions, it is essential to define the vertical axis ("Fail Rate"). In the left panels, the y-axis represents the proportion of a model's Top-1 predictions classified as outliers according to the Mahalanobis distance. In the right panels, it denotes the failure rate derived from the distributional Hotelling's $T^{2}$ test (Top-5 predictions).

A critical component for evaluating these metrics is the horizontal dashed line present across all panels. This line denotes the Human Fail Rate (derived as $1 - \text{Human Consistency}$ via the Leave-One-Out cross-validation detailed in Section~\ref{sec: Consistencia_humana}). Because human color association inherently contains semantic noise and subjective variance, an absolute error rate of zero is an unrealistic expectation. Instead, this dashed line establishes the empirical threshold of natural human disagreement. Consequently, for a model to be considered perfectly aligned with human perception, the bulk of its error distribution (the "belly" of the violin plot) should ideally center around or fall below this dashed empirical bound.

Regarding architectural families (Figure~\ref{fig:cuadricula_violines_completa}, top row), the Mahalanobis distance analysis reveals significant variability in how different structures anchor semantic concepts in the chromatic space. While classical architectures like the OpenAI family and compact models like MobileCLIP exhibit tighter distributions and lower median distances, newer architectures such as
CoCa and the EVA family display prominent long-tail distributions, indicating severe outliers for specific semantic concepts. This trend is corroborated by the Hotelling’s $T^2$ test. Although models like SigLIP achieve a relatively balanced alignment, a substantial portion of the distributions across all families remains well above the human baseline.

When grouped by pre-training dataset origin (Figure~\ref{fig:cuadricula_violines_completa}, bottom row), the data highlights the impact of corpus curation on perceptual alignment. Models trained on massive, uncurated scrapes like the Pioneer datasets (YFCC/CC) exhibit the highest dispersion and error rates. In contrast, models trained on highly curated or proprietary datasets (such as Owners/OpenAI and DataComp/DFN) demonstrate noticeably lower median error rates and more compact distributions. This suggests that the sheer volume of training data (as seen in LAION) is not sufficient to guarantee human-like color perception; data quality and rigorous filtering play a fundamental role in mitigating severe semantic misalignments.

To illustrate this global divergence with concrete instances, Table~\ref{tab:ejemplos} presents word-level examples of highly aligned and severely misaligned concepts across all models. For the highly aligned concepts, rather than exclusively selecting exact coordinate matches (which are comprehensively detailed in Appendix~\ref{app: Word_color}), we have curated examples that reflect the well-documented memory color phenomenon described by \citet{witzel_perception} and \citet{hansen_memory}. Specifically, humans do not simply recall the literal physical color of an object, but rather a prototypical, often more saturated or focal representation of it. The models that achieve high alignment successfully replicate this cognitive bias, converging with humans on these idealized representations for strong physical referents like \texttt{BANANA}, \texttt{PUMPKIN} or \texttt{PIG}.  

\begin{table*}[!h]
    \centering
    \caption{Word-level examples of highly aligned (left) and severely misaligned (right) concepts. The most frequent Human and Model predictions are broken down by their discrete game board coordinate and their mapped macro-color category.}
    \label{tab:ejemplos}
    \resizebox{\textwidth}{!}{
    \begin{tabular}{@{} l c l l l l || l c l l l l @{}}
        \toprule
        \multicolumn{6}{@{}c||}{\textbf{Highly Aligned Concepts (Low Error)}} & \multicolumn{6}{c@{}}{\textbf{Severely Misaligned Concepts (High Error)}} \\
        \cmidrule(r){1-6} \cmidrule(l){7-12}
        \textbf{Word} & \textbf{Hum. Cons.} & \textbf{H-Coord} & \textbf{H-Color} & \textbf{M-Coord} & \textbf{M-Color} & \textbf{Word} & \textbf{Hum. Cons.} & \textbf{H-Coord} & \textbf{H-Color} & \textbf{M-Coord} & \textbf{M-Color} \\
        \midrule
        BANANA  & 0.90 & \cellcolor[RGB]{238,198,69} I4   & Yellow & \cellcolor[RGB]{238,220,101} J7  & Yellow & APPLE       & 0.89 & \cellcolor[RGB]{194,54,46} A13  & Red    & \cellcolor[RGB]{191,193,194} I18 & Blue \\
        PUMPKIN & 0.86 & \cellcolor[RGB]{218,131,42} E4   & Orange & \cellcolor[RGB]{223,136,44} E3   & Orange & EGG         & 0.90 & \cellcolor[RGB]{213,114,50} E6  & Orange & \cellcolor[RGB]{238,220,101} J7  & Yellow \\
        PIG     & 0.89 & \cellcolor[RGB]{218,127,144} F15 & Pink   & \cellcolor[RGB]{218,165,158} H15 & Pink   & AXOLOTL     & 0.89 & \cellcolor[RGB]{222,144,155} G15& Pink   & \cellcolor[RGB]{191,193,194} I18 & Blue \\
        TIGER   & 0.86 & \cellcolor[RGB]{213,114,50} E6   & Orange & \cellcolor[RGB]{223,136,44} E3   & Orange & TOAD        & 0.80 & \cellcolor[RGB]{65,118,66} P10  & Green  & \cellcolor[RGB]{191,193,194} I18 & Blue \\
        BASIL   & 0.88 & \cellcolor[RGB]{78,140,69} O12   & Green  & \cellcolor[RGB]{145,178,120} L13 & Green  & LAVENDER    & 0.90 & \cellcolor[RGB]{145,64,120} B23 & Purple & \cellcolor[RGB]{142,131,158} H21 & Blue \\
        WOOD    & 0.85 & \cellcolor[RGB]{115,71,37} B1    & Brown  & \cellcolor[RGB]{143,92,43} C1    & Brown  & WHEAT       & 0.91 & \cellcolor[RGB]{230,162,42} F2  & Orange & \cellcolor[RGB]{207,190,146} I13 & Green \\
        EMERALD & 0.86 & \cellcolor[RGB]{116,167,70} N9   & Green  & \cellcolor[RGB]{63,120,66} P11   & Green  & RUBY        & 0.85 & \cellcolor[RGB]{196,61,46} B8   & Red    & \cellcolor[RGB]{197,53,86} A17   & Pink \\
        GOLD    & 0.94 & \cellcolor[RGB]{230,186,38} H1   & Yellow & \cellcolor[RGB]{221,168,33} G1   & Yellow & TILE        & 0.86 & \cellcolor[RGB]{143,48,41} A7   & Brown    & \cellcolor[RGB]{192,196,189} I17 & Blue \\
        BLOOD   & 0.89 & \cellcolor[RGB]{165,53,45} A8    & Red    & \cellcolor[RGB]{130,45,39} A6    & Red    & SAND        & 0.90 & \cellcolor[RGB]{232,172,88} G5  & Yellow & \cellcolor[RGB]{207,190,146} I13 & Green \\
        SUNSET  & 0.87 & \cellcolor[RGB]{205,95,49} D8    & Red    & \cellcolor[RGB]{235,159,109} H10 & Orange & THUNDERBOLT & 0.82 & \cellcolor[RGB]{234,219,67} K2  & Yellow & \cellcolor[RGB]{191,193,194} I18 & Blue \\
        AUTUMN  & 0.92 & \cellcolor[RGB]{128,70,41} B2    & Brown  & \cellcolor[RGB]{191,112,45} D3   & Orange & RAGE        & 0.85 & \cellcolor[RGB]{194,54,46} A13  & Red    & \cellcolor[RGB]{191,193,194} I18 & Blue \\
        ECOLOGY & 0.87 & \cellcolor[RGB]{77,150,66} O14   & Green  & \cellcolor[RGB]{126,173,68} N8   & Green  & WINTER      & 0.90 & \cellcolor[RGB]{96,164,187} L24 & Blue   & \cellcolor[RGB]{191,193,194} I18 & Blue \\
        BARBIE  & 0.84 & \cellcolor[RGB]{209,65,110} C16  & Pink   & \cellcolor[RGB]{204,68,122} C17  & Pink   & FEMINISM    & 0.85 & \cellcolor[RGB]{97,59,117} A30  & Purple & \cellcolor[RGB]{199,54,115} B18  & Pink \\
        HULK    & 0.75 & \cellcolor[RGB]{65,118,66} P10   & Green  & \cellcolor[RGB]{126,173,68} N8   & Green  & (LILO AND) STITCH      & 0.90 & \cellcolor[RGB]{81,147,175} M26 & Blue   & \cellcolor[RGB]{191,193,194} I18 & Blue \\
        \bottomrule
    \end{tabular}
    }
\end{table*}

Conversely, an analysis of the severely misaligned terms reveals two distinct modes of failure. The first type consists of semantic misclassifications, where the model selects an entirely incorrect color category, such as predicting green for \texttt{WHEAT}, pink for \texttt{RUBY}, or pink for \texttt{FEMINISM}. The second type of failure is characterized by an uncertainty collapse. In these cases, rather than making a misguided semantic association, the models systematically default to a specific blue coordinate (I18), completely disregarding the concept's actual visual properties or focal color (e.g., \texttt{APPLE}, \texttt{AXOLOTL}, or \texttt{TOAD}).

Ultimately, the most striking finding from this global overview is the generalized difficulty that CVLMs face in matching human consistency. The bulk of the probability density for almost every model family and dataset lies above the human failure rate. This statistically confirms a persistent human-model divergence: while CVLMs can successfully approximate broad color categories, they systematically fail to capture the nuanced, localized consensus of human memory color.

\subsection{Semantic Category Analysis}
\label{sec: results_semantic}

\begin{figure*}[h]
    \centering
    \includegraphics[width=1\linewidth]{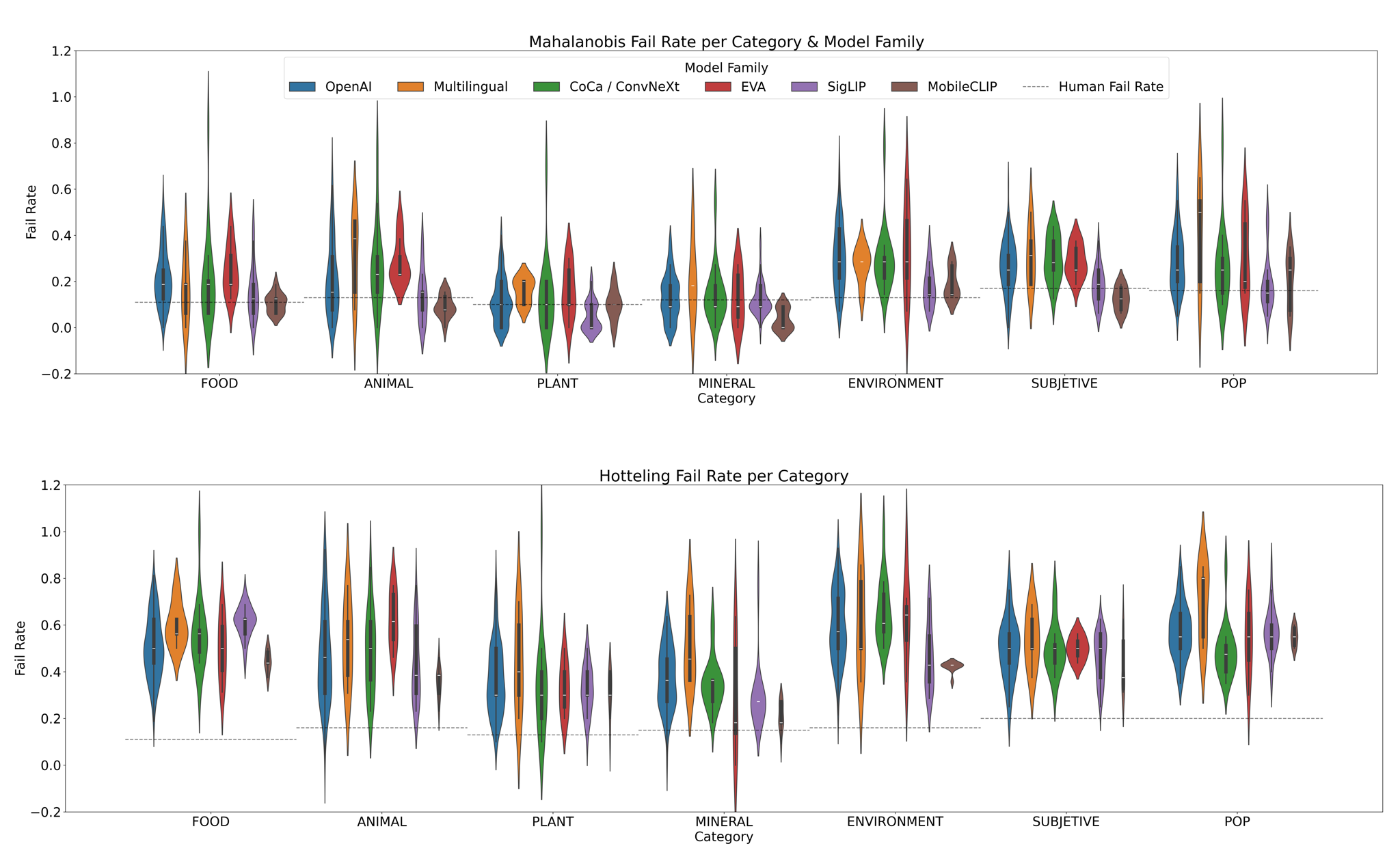}
    
    \caption{Fine-grained analysis of failure rates across semantic categories, grouped by architectural family. The graph above shows the Mahalanobis failure rate, whilst the graph below shows the results of the Hotelling $T^2$ test. Each violin plot groups together the performance of all model control points belonging to the same architectural family, relative to the human benchmark (horizontal dotted line).}
    \label{fig:ViolinCategoriaModelo}
\end{figure*}

To further isolate the drivers of perceptual misalignment, Figure \ref{fig:ViolinCategoriaModelo} presents a categorical breakdown of model performance. This granular view confirms that semantic difficulty is not uniform; models exhibit significantly higher robustness when grounding concrete entities—such as \textit{Food}, \textit{Animal}, and \textit{Plant}—compared to abstract or socially constructed domains like \textit{Subjective} and \textit{Pop Culture}.

A notable observation from the Mahalanobis distribution in Figure~\ref{fig:ViolinCategoriaModelo} is the superior consistency of the MobileCLIP family across most categories. While other architectures like OpenAI or Multilingual models exhibit wider variance (indicated by elongated violin shapes), MobileCLIP maintains a consistently narrower distribution, often approaching or even dipping below the human baseline in categories such as \textit{Plant} and \textit{Mineral}. This suggests that the architectural optimizations in MobileCLIP not only reduce the failure rate but also compress the uncertainty cloud of the model, leading to more predictable and human-aligned color associations.

In contrast, the \textit{Subjective} and \textit{Pop Culture} categories present a systematic challenge regardless of the architecture. The high median failure rates and broad distributions observed in Figure~\ref{fig:ViolinCategoriaModelo} indicate that when models lack a shared physical referent ("ground truth"), they do not converge on a single cultural consensus. Instead, the models likely sample from the fragmented "long-tail" of web data, reflecting the cultural biases and semantic ambiguities present in their respective pre-training corpora. These results suggest that the misalignment is not merely an architectural deficiency, but a reflection of the inherent statistical noise when grounding abstract concepts in Contrastive Vision-Language Models.

\begin{figure*}[h]
    \centering
    \includegraphics[width=1\linewidth]{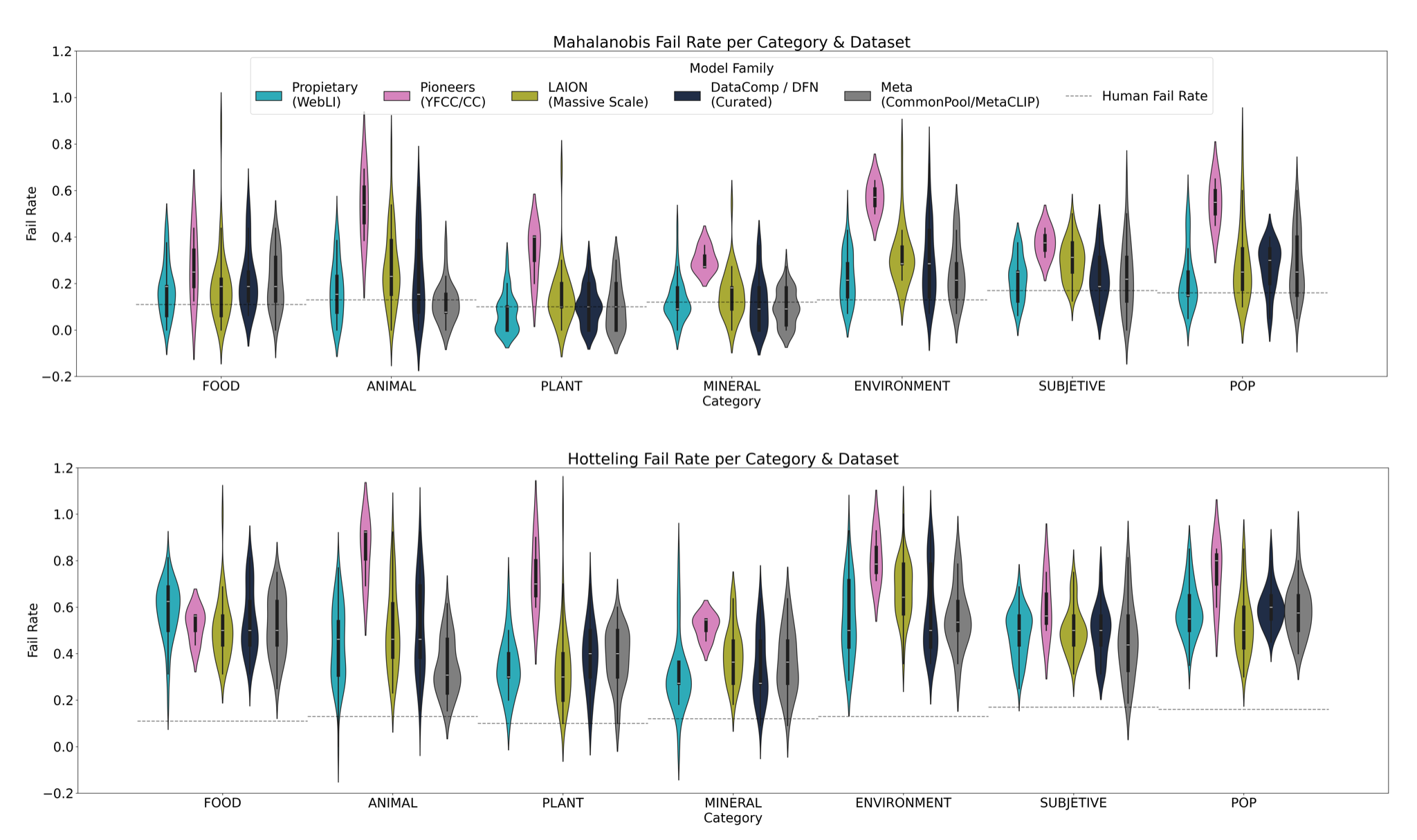}
    
    \caption{Fine-grained analysis of failure rates across semantic categories, grouped by pre-training dataset origin. The graph above shows the Mahalanobis failure rate, whilst the graph below shows the results of the Hotelling $T^{2}$ test. Each violin plot aggregates the performance of all model checkpoints trained on the corresponding dataset family, relative to the human benchmark (horizontal dotted line).}
    \label{fig:ViolinCategoriaDataset}
\end{figure*}

To complement the architectural perspective, Figure~\ref{fig:ViolinCategoriaDataset} extends this categorical analysis by grouping the models according to their pre-training dataset origin. This visualization corroborates that the gradient of difficulty—from concrete physical referents to abstract concepts—is fundamentally a data-driven phenomenon. In concrete categories such as Food, Animal, and Plant, models trained on highly curated or proprietary datasets (e.g., DataComp/DFN and WebLI) consistently exhibit tighter error distributions and lower medians that closely track the human baseline. Conversely, models relying on massive, uncurated web crawls (like LAION or the Pioneer datasets) present elongated long-tail distributions, indicating severe and frequent chromatic hallucinations even for well-defined objects.

However, when evaluating abstract domains (Subjective and Pop Culture), Figure~\ref{fig:ViolinCategoriaDataset} reveals a critical limitation: the variance explodes across all dataset families. Even the most rigorously curated corpora fail to converge on a single cultural consensus, confirming that while data quality can significantly refine physical grounding, it cannot easily resolve the inherent semantic ambiguity of concepts lacking a universal visual referent

To systematically quantify this categorical divergence, Table 3 presents the mean failure rates and errors aggregated by semantic category and architectural family. These values were derived through a rigorous three-step aggregation process: first, we calculated the individual failure rate for each model on every single word; second, we averaged these word-level scores to obtain a consolidated failure rate per model for each semantic category; and finally, we grouped the models by their architectural family to calculate the overall mean failure rate.

As observed in these results, the semantic domain strictly dictates the severity of the perceptual misalignment, directly reflecting the gradient from concrete "Memory Colors" to abstract concepts. First, the Human Fail Rate validates this gradient from concrete "Memory Colors" to abstract concepts. First, the \textit{Human Fail Rate} validates this gradient: human inconsistency increases from 0.110 in the \textit{Food} category to 0.200 in \textit{Pop Culture}, confirming that cultural and subjective concepts inherently possess a wider, more ambiguous ground truth.

\begin{table*}[!h]
    \centering
    \caption{Failure ratios per semantic category and architectural family. The table presents the mean failure rate and mean error evaluated under the Mahalanobis distance and the Hotelling's $T^2$ test ($2\sigma$ confidence threshold). The \textit{Human Fail Rate} column denotes the empirical baseline error per category, calculated as 1 minus human consistency via Leave-One-Out validation. The values highlighted in bold indicate the models that most closely resemble the Human Fail Rate.}
    \label{tab:tab_mah_hot}
    \resizebox{\textwidth}{!}{
    \begin{tabular}{@{}l c l c c c c | l c l c c c c@{}}
        \toprule
        \multirow{2}{*}{\textbf{Category}} & \textbf{Human} & \multirow{2}{*}{\textbf{Model Family}} & \multicolumn{2}{c}{\textbf{Mahalanobis}} & \multicolumn{2}{c|}{\textbf{Hotelling's $T^2$ ($2\sigma$)}} & \multirow{2}{*}{\textbf{Category}} & \textbf{Human} & \multirow{2}{*}{\textbf{Model Family}} & \multicolumn{2}{c}{\textbf{Mahalanobis}} & \multicolumn{2}{c}{\textbf{Hotelling's $T^2$ ($2\sigma$)}} \\
        \cmidrule(lr){4-5} \cmidrule(lr){6-7} \cmidrule(lr){11-12} \cmidrule(l){13-14}
        & \textbf{Fail Rate} & & \textbf{Mean} & \textbf{Mean Err} & \textbf{Mean} & \textbf{Mean Err} & & \textbf{Fail Rate} & & \textbf{Mean} & \textbf{Mean Err} & \textbf{Mean} & \textbf{Mean Err} \\
        \midrule
        \multirow{6}{*}{\textbf{Food}} 
        & \multirow{6}{*}{\textit{0.11}} 
        & CoCa / ConvNeXt          & 0.21 & 0.05 & 0.55 & 0.04 & \multirow{6}{*}{\textbf{Environment}} & \multirow{6}{*}{\textit{0.16}} & CoCa / ConvNeXt   & 0.30 & 0.04 & 0.67 & 0.03 \\
        & & EVA                    & 0.25 & 0.04 & 0.50 & 0.05 & & & EVA             & 0.34 & 0.08 & 0.65 & 0.06 \\
        & & SigLIP                 & 0.15 & 0.02 & 0.635 & 0.013 & & & \textbf{SigLIP}          & \textbf{0.176} & 0.016 & 0.49 & 0.03 \\
        & & \textbf{MobileCLIP}    & \textbf{0.106} & 0.013 & \textbf{0.444} & 0.015 & & & MobileCLIP      & 0.19 & 0.02 & \textbf{0.421} & 0.007 \\
        & & Multilingual           & 0.16 & 0.06 & 0.60 & 0.04 & & & Multilingual    & 0.27 & 0.03 & 0.61 & 0.10 \\
        & & OpenAI                 & 0.215 & 0.013 & 0.516 & 0.014 & & & OpenAI          & 0.305 & 0.015 & 0.622 & 0.016 \\
        \cmidrule(r){1-7} \cmidrule(l){8-14}
        \multirow{6}{*}{\textbf{Animal}} 
        & \multirow{6}{*}{\textit{0.16}} 
        & CoCa / ConvNeXt          & 0.26 & 0.05 & 0.47 & 0.04 & \multirow{6}{*}{\textbf{Subjective}} & \multirow{6}{*}{\textit{0.19}} & CoCa / ConvNeXt   & 0.30 & 0.02 & 0.50 & 0.03 \\
        & & EVA                    & 0.29 & 0.04 & 0.63 & 0.05 & & & EVA             & 0.29 & 0.03 & 0.509 & 0.016 \\
        & & \textbf{SigLIP}        & \textbf{0.13} & 0.02 & 0.43 & 0.03 & & & \textbf{SigLIP}          & \textbf{0.181} & 0.014 & 0.463 & 0.018 \\
        & & MobileCLIP             & 0.092 & 0.015 & \textbf{0.35} & 0.02 & & & MobileCLIP      & 0.125 & 0.016 & \textbf{0.43} & 0.04 \\
        & & Multilingual           & 0.31 & 0.08 & 0.52 & 0.08 & & & Multilingual    & 0.31 & 0.06 & 0.53 & 0.06 \\
        & & OpenAI                 & 0.213 & 0.017 & 0.47 & 0.02 & & & OpenAI          & 0.263 & 0.012 & 0.495 & 0.014 \\
        \cmidrule(r){1-7} \cmidrule(l){8-14}
        \multirow{6}{*}{\textbf{Plant}} 
        & \multirow{6}{*}{\textit{0.13}} 
        & \textbf{CoCa / ConvNeXt} & \textbf{0.14} & 0.04 & 0.36 & 0.05 & \multirow{6}{*}{\textbf{Pop Culture}} & \multirow{6}{*}{\textit{0.20}} & CoCa / ConvNeXt   & 0.27 & 0.04 & \textbf{0.51} & 0.03 \\
        & & EVA                    & 0.16 & 0.04 & 0.36 & 0.04 & & & EVA             & 0.31 & 0.06 & 0.55 & 0.06 \\
        & & SigLIP                 & 0.050 & 0.011 & 0.330 & 0.019 & & & SigLIP          & 0.19 & 0.02 & 0.572 & 0.019 \\
        & & MobileCLIP             & 0.10 & 0.02 & \textbf{0.32} & 0.04 & & & \textbf{MobileCLIP}      & \textbf{0.21} & 0.04 & 0.560 & 0.012 \\
        & & Multilingual           & 0.16 & 0.02 & 0.42 & 0.09 & & & Multilingual    & 0.41 & 0.10 & 0.71 & 0.06 \\
        & & OpenAI                 & 0.116 & 0.010 & 0.387 & 0.016 & & & OpenAI          & 0.287 & 0.014 & 0.590 & 0.012 \\
        \cmidrule(r){1-7} \cmidrule(l){8-14}
        \multirow{6}{*}{\textbf{Mineral}} 
        & \multirow{6}{*}{\textit{0.15}} 
        & \textbf{CoCa / ConvNeXt} & \textbf{0.15} & 0.03 & 0.37 & 0.03 & & & & & & & \\
        & & EVA                    & 0.13 & 0.04 & 0.31 & 0.09 & & & & & & & \\
        & & SigLIP                 & 0.124 & 0.013 & 0.29 & 0.03 & & & & & & & \\
        & & MobileCLIP             & 0.036 & 0.015 & \textbf{0.209} & 0.019 & & & & & & & \\
        & & Multilingual           & 0.20 & 0.07 & 0.49 & 0.06 & & & & & & & \\
        & & OpenAI                 & 0.132 & 0.010 & 0.379 & 0.014 & & & & & & & \\
        \bottomrule
    \end{tabular}
    }
\end{table*}

Remarkably, when evaluating the Top-1 prediction via the Mahalanobis distance, highly optimized architectures occasionally surpass human consistency in concrete domains. For instance, in the \textit{Plant} category, the SigLIP family achieves a failure rate of just 0.030 (compared to the 0.130 human baseline), and in \textit{Mineral}, MobileCLIP reaches 0.036 (versus 0.150 for humans). This visual evidence suggests that for concepts tightly bound to a physical reality, CVLMs can pinpoint the theoretical centroid with extreme precision.

However, a critical collapse is revealed when assessing the distributional alignment (Top-5) through the Hotelling's $T^2$ test. Even for models that achieve near-perfect Mahalanobis scores, the Hotelling failure rates escalate dramatically across all categories, frequently ranging between 0.400 and 0.700. This indicates a profound limitation: while CVLMs can accurately identify a prototypical color coordinate, their broader "confidence cloud" fails to capture the nuanced, secondary color associations that humans naturally make. This divergence is exacerbated in abstract categories like \textit{Pop Culture} and \textit{Subjective}, where models—particularly Multilingual architectures (0.700 Hotelling failure rate in \textit{Pop Culture})—lack a universal physical referent and consequently sample from fragmented, noisy cultural biases present in their pre-training data.

\subsection{The Impact of Uncertainty}
\label{sec:results_collapse}

To rigorously contextualize the errors observed in valid semantic categories, we introduced a "Nonsense Baseline" to probe the models' behavior under total semantic uncertainty. We queried the models with a 16-item vocabulary of semantically void inputs, including numbers, stop words, pseudowords, and random alphanumeric strings.

\begin{figure*}[!h]
	\centering
	\includegraphics[width=1\linewidth]{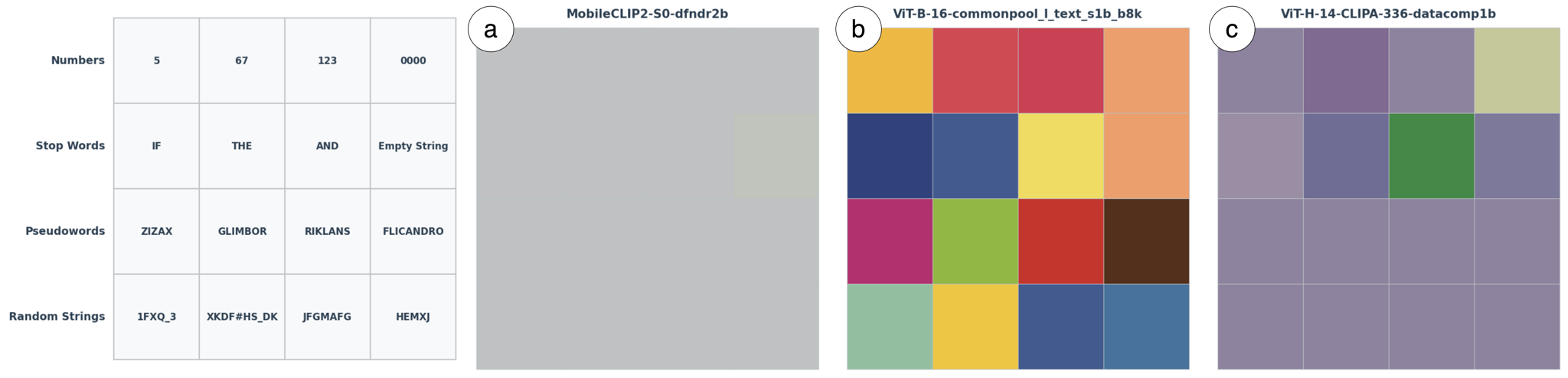}
	\caption{Typology of model responses to the "Nonsense Baseline" (semantically void inputs). The leftmost panel defines the 16-word control vocabulary. We identify four distinct behavioral modes under total uncertainty: (a)~Neutral Collapse, where the model defaults to the most achromatic coordinate available on the board (I18), acting as a logical neutral fallback; (b)~High Entropy, where the model samples randomly across the chromatic space, showing no learned prior; (c)~Specific Color Collapse, where the model systematically defaults to a strong focal color; and (d)~Selective Collapse, where the model exhibits a systematic bias exclusively triggered by specific linguistic categories (e.g., stop words). These biased collapses (c and d) highlight a critical risk of false positives, where a model might appear semantically aligned with a word simply because the target color matches its structural or dataset-induced prior.}
	\label{fig:TipoColapso}
\end{figure*}

Analyzing the Top-1 color predictions for these inputs reveals that Contrastive Vision-Language Models do not handle uncertainty uniformly. To systematically classify the models' uncertainty responses, we established a quantitative heuristic based on the macro-color categorization defined in Section~\ref{sec: color_naming}. First, we evaluated the predictions at the row level (i.e., within each of the four linguistic categories). A single row is considered to "collapse" if at least three of its four cells share the exact same macro-color category. Based on this row-level agreement, as illustrated in Figure~\ref{fig:TipoColapso}, we assigned the models to one of three distinct behavioral modes:

(a)~Neutral Collapse: A model is assigned to this category if three or more of its rows collapse into coordinate I18. While technically located within the blue region due to the discrete layout of the Hues and Cues board, I18 represents the most achromatic, grayish cell available. In the absence of semantic meaning, defaulting to a neutral, desaturated tone serves as a logical and acceptable fallback mechanism.

(b)~High Entropy (No Collapse): Models that do not meet the criteria for a systematic global collapse (i.e., fewer than three rows collapse in total, or they collapse into different, unrelated colors) are assigned to this category. These architectures exhibit no specific chromatic preference when faced with void inputs, instead sampling randomly across the board, indicating the absence of a strong visual prior for meaningless tokens.

(c)~Specific Color Collapse: A model is assigned to this category if three or more of its rows collapse into the exact same non-neutral focal color (e.g., blues, yellows, or reds). This behavior reveals a significant underlying bias across the entire nonsense baseline, stemming either from the architectural transformation of the visual input or from imbalances within the pre-training dataset.


This latter mode of collapse (Specific) acts as a warning sign when evaluating CVLM alignment, as they introduce a high risk of false positives. If a model's intrinsic prior happens to align with the target color of a specific valid concept, the model may falsely appear to possess semantic understanding, when in reality it is merely defaulting to its learned bias. Consequently, establishing this "Nonsense Baseline" is essential for validating true perceptual grounding, ensuring that correct predictions stem from semantic comprehension rather than statistical coincidence. A comprehensive visualization of these collapses for all 162 evaluated models is provided in Appendix~\ref{app: Colapso}.

Table~\ref{tab:collapse_distribution} quantifies the distribution of these three behavioral modes across the 162 evaluated models. According to our analysis, a combined 58.0\% of the architectures exhibit a valid and robust response to total semantic uncertainty. This encompasses both the 35.2\% of models that display high-entropy behavior—distributing their predictions randomly in the absence of a visual prior—and the 22.8\% that adopt a Neutral Collapse, logically defaulting to the most achromatic coordinate available (I18). However, the remaining 42.0\% of the models exhibit biased behaviors that fail to handle ambiguity safely, either through a Specific Color Collapse (17.3\%) or a Selective Collapse (24.7\%). This quantitative breakdown emphasizes that the risk of false positives—where models systematically hallucinate strong focal colors that might coincidentally masquerade as semantic understanding—is a widespread structural characteristic affecting a substantial portion of current Contrastive Vision-Language Models.

\begin{table}[h]
    \centering
    \caption{Distribution of the 162 evaluated Contrastive Vision-Language Models across the three behavioral modes identified during the Nonsense Baseline test.}
    \label{tab:collapse_distribution}
    \begin{tabular}{@{} l c c @{}}
        \toprule
        \textbf{Behavioral Mode} & \textbf{Models} & \textbf{Percentage (\%)} \\
        \midrule
        (a) Neutral Collapse & 45 & 27.78 \\
        (b) High Entropy (No Collapse) & 73 & 45.06 \\
        (c) Specific Color Collapse & 44 & 27.16 \\
        \midrule
        \textbf{Total} & \textbf{162} & \textbf{100.0} \\
        \bottomrule
    \end{tabular}
\end{table}

\section{Discussion and Final Remarks}

This work introduces a gamified evaluation framework based on the board game Hues and Cues to probe the fine-grained semantic color alignment of 162 Contrastive Vision-Language Models (CVLMs) against an empirical human baseline. By translating discrete game choices into the standard CIE xy chromaticity diagram, our framework bypasses the issues of infinite continuous color spaces while capturing the cognitive, culturally anchored nuances of human memory color. The resulting findings yield several key insights into the architectural limits, dataset influences, and semantic boundaries of modern visual-language models.  

\subsection{Real-World Grounds vs. Subjective Abstractions}

Our results show a stark contrast in model alignment between concrete physical referents and abstract domains. In concrete categories like Food, Animal, and Plant, highly optimized models (such as those in the SigLIP and MobileCLIP families) frequently match or even exceed human consistency when looking at point-wise predictions (Top-1 Mahalanobis distance). For concepts like \texttt{BANANA}, \texttt{PUMPKIN}, or \texttt{PIG}, the models successfully replicate the human cognitive bias toward idealized, saturated memory colors rather than literal physical hues. This indicates that contrastive pre-training effectively captures highly persistent physical statistical associations present on the web.  However, this alignment collapses catastrophically when evaluating abstract or socially constructed concepts (Subjective and Pop Culture). Because concepts like \texttt{DANGER}, \texttt{HOPE}, \texttt{FEMINISM}, or \texttt{TRACTOR} lack a single, uniform physical referent in nature, humans rely on shared cultural consensus or symbolic metaphors. CVLMs systematically diverge from this human consensus. Instead of tracking the localized cultural anchors shared by biological observers, the models sample from fragmented, high-variance representations across web-scraped data, resulting in massive distributional misalignments.  

\subsection{Failure Modes: Misclassification and Uncertainty Collapse}
A deeper examination of severely misaligned concepts reveals two distinct structural failure modes:  
\begin{itemize}
\item \textbf{Semantic Misclassification:} This occurs when a model actively links a concept to an entirely incorrect color category, such as predicting green for \texttt{WHEAT}, pink for \texttt{RUBY}, or green for \texttt{SAND}. This points to a clear failure in semantic grounding where text-to-image embeddings cross major categorical color boundaries.  

\item \textbf{Uncertainty Collapse:} Rather than mapping a concept to an incorrect but deliberate focal color, models confronted with semantic ambiguity frequently collapse into a default chromatic coordinate (I18, which corresponds to an achromatic gray-blue). This behavior is highly similar to the distribution collapses and class-biases observed in broad VLM hallucination studies.  Our fine-grained distributional analysis using Hotelling's $T^2$ test confirms that even when a model's primary guess is accurate, its broader "confidence cloud" (Top-5 predictions) fails to mirror the rich, multi-modal secondary color distributions that humans naturally make.  
\end{itemize}

\subsection{The Nonsense Baseline and the Risk of False Positives}

By evaluating semantically void prompts (such as stop words, empty strings, and random characters), we confirmed that 27.16\% of the evaluated models suffer from structurally or dataset-induced biases, showing Specific Color Collapses. Models displaying these behaviors systematically spit out strong focal colors for entirely meaningless tokens.  This discovery introduces a major cautionary tale for AI evaluation: a model might appear to "understand" the color of a valid concept (such as a model defaulting to blue for \texttt{WINTER} or \texttt{STITCH}) simply because the concept's target color coincidentally aligns with the model's structural or dataset-induced uncertainty prior. Therefore, incorporating a "Nonsense Baseline" is essential for future perceptual benchmarking to differentiate genuine semantic understanding from artifactual statistical coincidence.  

\subsection{Curation Over Scale}

Finally, our global evaluation sheds light on the ongoing "data scale vs. data quality" debate in machine learning. While newer, larger architectures like CoCa and the EVA family display prominent long-tail error distributions, compact architectures with highly curated pre-training data show significantly tighter, more human-aligned behavior. Massive, uncurated web crawls (e.g., the YFCC or early LAION pipelines) introduce tremendous statistical noise and overrepresentations of specific structural or corporate hues. In contrast, models pre-trained on rigorously filtered datasets (such as DataComp, DFN, or OpenAI's proprietary sets) demonstrate that clean, high-quality data curation is a far more effective tool for building structurally balanced and human-aligned visual spaces than sheer data volume or model parameters. 
This is consistent with previous reports on the relevance of the statistics of the training set so that human-like color discrimination properties emerge~\cite{Hernandez24}.

\subsection{Limitations and Future Work}

While our framework successfully exposes latent cognitive deficiencies in CVLMs, a few limitations should be noted. First, human data collection from the 325 participants was crowdsourced remotely on personal devices. Although semantic color categorization is largely robust to minor screen variations —as suggested by the statistical similarity with color constancy under illumination change Appendix~\ref{app:Adaptation}—  absolute colorimetric precision could be further validated under standardized laboratory lighting conditions. Second, our evaluation focused on contrastive, retrieval-based models. Future work should expand this gamified protocol to evaluate auto-regressive large vision-language models (e.g., the GPT-V or LLaVA families) to see if generative visual spaces display the same uncertainty collapses and cultural misalignments.

\section*{Data and code availability}
The dataset of human color associations, the evaluation results for the 162 CVLMs, and the source code required to reproduce the metrics and figures presented in this study are publicly available in our GitHub repository: \url{https://github.com/Rietta5/HuesAndCues}.

\section*{Acknowledgements}
The work was partially funded by the Spanish Government and the EU under the MCIN/AEI/FEDER/UE Grant PID2023-152133NB-I00, the CIACIF/2023/223 from Spanish GVA, the FPU21/02256 from the Spanish MIU and by the BBVA Foundations of Science program in Maths, Stats, Comp. Sci. and AI, grant VIS4NN: \emph{Vision Science for Artificial Neural Networks}.

\appendix

\section{Complete List of Evaluated
Contrastive Vision-Language Models}
\label{app:ListadoCompleto}

This appendix provides the complete inventory of the Contrastive Vision-Language Model checkpoints evaluated in this study. The compilation encompasses all architectural families—including classical models, SigLIP, EVA, MobileCLIP, CoCa, and ConvNeXt—and spans the diverse pre-training corpora discussed in Section~\ref{sec:ModelBenchmarking}. Additionally, the table details the specific uncertainty collapse behavior (Neutral Collapse, High Entropy, or Specific Color Collapse) exhibited by each model when subjected to the Nonsense Baseline experiment (Section~\ref{sec: colapso}). The models are listed alphabetically for reference in Table~\ref{tab:all_models_list}.

\begin{table*}[!h]
\centering
\caption{Complete list of the evaluated Contrastive Vision-Language Models with their assigned identifier used in Figure~\ref{fig:GlobalColapso}. \emph{Row Colors} lists the collapse colour of each mosaic row in order (\emph{---}\ if no collapse). \emph{Cat.}:\ A)~Neutral Collapse, B)~High Entropy, C)~Specific Color Collapse.}
\label{tab:all_models_list}
\begin{adjustbox}{max width=\textwidth}
\small
\begin{tabular}{@{} r l l c  r l l c @{}}
\toprule
\textbf{\#} & \textbf{Model} & \textbf{Row Colors} & \textbf{Cat.} & \textbf{\#} & \textbf{Model} & \textbf{Row Colors} & \textbf{Cat.} \\
\midrule
1 & EVA01-g-14-laion400m\_s11b\_b41k & pink / yellow / — / — & B & 2 & EVA01-g-14-plus-merged2b\_s11b\_b114k & yellow / — / blue / achromatic & B \\
3 & EVA02-B-16-merged2b\_s8b\_b131k & achromatic / — / achromatic / achromatic & A & 4 & EVA02-E-14-plus-laion2b\_s9b\_b144k & — / — / blue / — & B \\
5 & EVA02-L-14-336-merged2b\_s6b\_b61k & blue / blue / — / — & C & 6 & EVA02-L-14-merged2b\_s4b\_b131k & brown / brown / blue / — & C \\
7 & MobileCLIP-B-datacompdr & orange / — / orange / — & C & 8 & MobileCLIP-B-datacompdr\_lt & — / — / — / — & B \\
9 & MobileCLIP-S1-datacompdr & yellow / yellow / yellow / — & C & 10 & MobileCLIP-S2-datacompdr & yellow / yellow / — / — & C \\
11 & MobileCLIP2-B-dfndr2b & achromatic / achromatic / achromatic / achromatic & A & 12 & MobileCLIP2-L-14-dfndr2b & achromatic / achromatic / achromatic / achromatic & A \\
13 & MobileCLIP2-S0-dfndr2b & achromatic / achromatic / achromatic / achromatic & A & 14 & MobileCLIP2-S2-dfndr2b & achromatic / achromatic / achromatic / achromatic & A \\
15 & MobileCLIP2-S3-dfndr2b & achromatic / achromatic / achromatic / achromatic & A & 16 & MobileCLIP2-S4-dfndr2b & achromatic / achromatic / achromatic / achromatic & A \\
17 & PE-Core-B-16-meta & green / green / green / green & C & 18 & PE-Core-L-14-336-meta & pink / pink / green / pink & C \\
19 & PE-Core-S-16-384-meta & achromatic / achromatic / — / achromatic & A & 20 & PE-Core-T-16-384-meta & green / achromatic / green / achromatic & C \\
21 & PE-Core-bigG-14-448-meta & pink / — / — / — & B & 22 & RN101-openai & — / — / — / — & B \\
23 & RN101-yfcc15m & — / — / — / — & B & 24 & RN50-cc12m & — / achromatic / — / achromatic & A \\
25 & RN50-openai & — / yellow / purple / — & B & 26 & RN50-yfcc15m & — / achromatic / achromatic / — & A \\
27 & RN50x16-openai & — / — / — / blue & B & 28 & RN50x4-openai & — / — / — / — & B \\
29 & RN50x64-openai & — / — / — / — & B & 30 & ViT-B-16-SigLIP-256-webli & achromatic / — / — / — & B \\
31 & ViT-B-16-SigLIP-384-webli & achromatic / — / — / — & B & 32 & ViT-B-16-SigLIP-512-webli & achromatic / — / — / — & B \\
33 & ViT-B-16-SigLIP-i18n-256-webli & — / green / — / achromatic & B & 34 & ViT-B-16-SigLIP-webli & achromatic / — / green / — & B \\
35 & ViT-B-16-SigLIP2-256-webli & achromatic / achromatic / achromatic / achromatic & A & 36 & ViT-B-16-SigLIP2-384-webli & achromatic / achromatic / achromatic / achromatic & A \\
37 & ViT-B-16-SigLIP2-512-webli & achromatic / achromatic / achromatic / achromatic & A & 38 & ViT-B-16-SigLIP2-webli & achromatic / achromatic / achromatic / achromatic & A \\
39 & ViT-B-16-commonpool\_l\_basic\_s1b\_b8k & — / — / blue / blue & C & 40 & ViT-B-16-commonpool\_l\_clip\_s1b\_b8k & — / — / achromatic / achromatic & A \\
41 & ViT-B-16-commonpool\_l\_image\_s1b\_b8k & red / — / — / — & B & 42 & ViT-B-16-commonpool\_l\_laion\_s1b\_b8k & — / achromatic / achromatic / — & A \\
43 & ViT-B-16-commonpool\_l\_s1b\_b8k & orange / — / — / — & B & 44 & ViT-B-16-commonpool\_l\_text\_s1b\_b8k & — / — / — / — & B \\
45 & ViT-B-16-datacomp\_l\_s1b\_b8k & — / achromatic / — / — & B & 46 & ViT-B-16-datacomp\_xl\_s13b\_b90k & — / — / — / — & B \\
47 & ViT-B-16-dfn2b & achromatic / achromatic / achromatic / achromatic & A & 48 & ViT-B-16-laion2b\_s34b\_b88k & — / — / — / achromatic & B \\
49 & ViT-B-16-laion400m\_e31 & pink / — / — / blue & B & 50 & ViT-B-16-laion400m\_e32 & pink / yellow / — / blue & B \\
51 & ViT-B-16-metaclip\_400m & — / — / achromatic / — & B & 52 & ViT-B-16-metaclip\_fullcc & pink / — / — / achromatic & B \\
53 & ViT-B-16-openai & — / yellow / — / — & B & 54 & ViT-B-16-plus-240-laion400m\_e31 & achromatic / — / blue / blue & C \\
55 & ViT-B-16-plus-240-laion400m\_e32 & achromatic / — / blue / blue & C & 56 & ViT-B-32-256-datacomp\_s34b\_b86k & — / blue / blue / — & C \\
57 & ViT-B-32-SigLIP2-256-webli & achromatic / achromatic / achromatic / achromatic & A & 58 & ViT-B-32-commonpool\_m\_basic\_s128m\_b4k & — / yellow / yellow / — & C \\
59 & ViT-B-32-commonpool\_m\_clip\_s128m\_b4k & blue / achromatic / — / — & B & 60 & ViT-B-32-commonpool\_m\_image\_s128m\_b4k & — / pink / pink / pink & C \\
61 & ViT-B-32-commonpool\_m\_laion\_s128m\_b4k & — / — / — / — & B & 62 & ViT-B-32-commonpool\_m\_s128m\_b4k & — / — / — / — & B \\
63 & ViT-B-32-commonpool\_m\_text\_s128m\_b4k & brown / — / — / — & B & 64 & ViT-B-32-commonpool\_s\_basic\_s13m\_b4k & — / — / — / blue & B \\
65 & ViT-B-32-commonpool\_s\_clip\_s13m\_b4k & green / — / green / — & C & 66 & ViT-B-32-commonpool\_s\_image\_s13m\_b4k & — / blue / — / — & B \\
67 & ViT-B-32-commonpool\_s\_laion\_s13m\_b4k & — / — / — / — & B & 68 & ViT-B-32-commonpool\_s\_s13m\_b4k & yellow / blue / — / — & B \\
69 & ViT-B-32-commonpool\_s\_text\_s13m\_b4k & blue / blue / blue / blue & C & 70 & ViT-B-32-datacomp\_m\_s128m\_b4k & blue / blue / — / blue & C \\
71 & ViT-B-32-datacomp\_s\_s13m\_b4k & — / blue / — / — & B & 72 & ViT-B-32-datacomp\_xl\_s13b\_b90k & — / achromatic / — / — & B \\
73 & ViT-B-32-laion2b\_e16 & pink / blue / blue / blue & C & 74 & ViT-B-32-laion2b\_s34b\_b79k & blue / blue / blue / blue & C \\
75 & ViT-B-32-laion400m\_e31 & — / pink / — / achromatic & B & 76 & ViT-B-32-laion400m\_e32 & pink / pink / — / achromatic & C \\
77 & ViT-B-32-metaclip\_400m & pink / achromatic / achromatic / — & A & 78 & ViT-B-32-metaclip\_fullcc & — / — / achromatic / achromatic & A \\
79 & ViT-B-32-openai & blue / blue / purple / — & C & 80 & ViT-H-14-378-dfn5b & achromatic / — / achromatic / achromatic & A \\
81 & ViT-H-14-CLIPA-336-datacomp1b & purple / — / purple / purple & C & 82 & ViT-H-14-CLIPA-336-laion2b & achromatic / achromatic / — / achromatic & A \\
83 & ViT-H-14-CLIPA-datacomp1b & purple / — / purple / purple & C & 84 & ViT-H-14-dfn5b & achromatic / achromatic / achromatic / achromatic & A \\
85 & ViT-H-14-laion2b\_s32b\_b79k & achromatic / achromatic / achromatic / achromatic & A & 86 & ViT-H-14-metaclip\_altogether & — / achromatic / blue / green & B \\
87 & ViT-H-14-metaclip\_fullcc & achromatic / achromatic / — / — & A & 88 & ViT-H-14-worldwide-378-metaclip2\_worldwide & pink / — / — / — & B \\
89 & ViT-H-14-worldwide-metaclip2\_worldwide & brown / — / — / — & B & 90 & ViT-L-14-336-openai & purple / purple / green / — & C \\
91 & ViT-L-14-CLIPA-336-datacomp1b & — / — / purple / purple & C & 92 & ViT-L-14-CLIPA-datacomp1b & purple / purple / purple / purple & C \\
93 & ViT-L-14-commonpool\_xl\_clip\_s13b\_b90k & — / achromatic / — / achromatic & A & 94 & ViT-L-14-commonpool\_xl\_laion\_s13b\_b90k & — / — / — / blue & B \\
95 & ViT-L-14-commonpool\_xl\_s13b\_b90k & — / — / yellow / — & B & 96 & ViT-L-14-datacomp\_xl\_s13b\_b90k & — / — / — / — & B \\
97 & ViT-L-14-dfn2b & achromatic / green / achromatic / achromatic & A & 98 & ViT-L-14-dfn2b\_s39b & — / — / achromatic / achromatic & A \\
99 & ViT-L-14-laion2b\_s32b\_b82k & achromatic / — / — / achromatic & A & 100 & ViT-L-14-laion400m\_e31 & red / — / blue / blue & C \\
101 & ViT-L-14-laion400m\_e32 & red / red / blue / blue & C & 102 & ViT-L-14-metaclip\_400m & brown / — / — / — & B \\
103 & ViT-L-14-metaclip\_fullcc & — / — / yellow / yellow & C & 104 & ViT-L-14-openai & — / purple / green / yellow & B \\
105 & ViT-L-16-SigLIP-256-webli & achromatic / purple / — / — & B & 106 & ViT-L-16-SigLIP-384-webli & achromatic / purple / — / — & B \\
107 & ViT-L-16-SigLIP2-256-webli & achromatic / — / achromatic / — & A & 108 & ViT-L-16-SigLIP2-384-webli & brown / — / achromatic / — & B \\
109 & ViT-L-16-SigLIP2-512-webli & achromatic / achromatic / achromatic / yellow & A & 110 & ViT-SO400M-14-SigLIP-378-webli & achromatic / purple / — / — & B \\
111 & ViT-SO400M-14-SigLIP-384-webli & achromatic / purple / — / — & B & 112 & ViT-SO400M-14-SigLIP-webli & achromatic / — / — / — & B \\
113 & ViT-SO400M-14-SigLIP2-378-webli & achromatic / achromatic / — / — & A & 114 & ViT-SO400M-14-SigLIP2-webli & achromatic / achromatic / achromatic / achromatic & A \\
115 & ViT-SO400M-16-SigLIP-i18n-256-webli & achromatic / — / — / — & B & 116 & ViT-SO400M-16-SigLIP2-256-webli & achromatic / achromatic / achromatic / achromatic & A \\
117 & ViT-SO400M-16-SigLIP2-384-webli & achromatic / achromatic / — / — & A & 118 & ViT-SO400M-16-SigLIP2-512-webli & achromatic / — / achromatic / — & A \\
119 & ViT-bigG-14-CLIPA-336-datacomp1b & purple / purple / purple / purple & C & 120 & ViT-bigG-14-CLIPA-datacomp1b & purple / purple / purple / purple & C \\
121 & ViT-bigG-14-laion2b\_s39b\_b160k & achromatic / — / — / — & B & 122 & ViT-bigG-14-metaclip\_fullcc & achromatic / blue / achromatic / — & A \\
123 & ViT-bigG-14-worldwide-378-metaclip2\_worldwide & achromatic / — / — / — & B & 124 & ViT-bigG-14-worldwide-metaclip2\_worldwide & — / — / — / — & B \\
125 & ViT-g-14-laion2b\_s12b\_b42k & achromatic / — / blue / blue & C & 126 & ViT-g-14-laion2b\_s34b\_b88k & — / blue / blue / blue & C \\
127 & ViT-gopt-16-SigLIP2-256-webli & — / achromatic / achromatic / — & A & 128 & ViT-gopt-16-SigLIP2-384-webli & achromatic / achromatic / achromatic / — & A \\
129 & ViTamin-B-LTT-datacomp1b & — / — / — / green & B & 130 & ViTamin-B-datacomp1b & — / blue / blue / — & C \\
131 & ViTamin-L-256-datacomp1b & — / — / — / green & B & 132 & ViTamin-L-datacomp1b & blue / — / — / — & B \\
133 & ViTamin-L2-256-datacomp1b & purple / purple / blue / — & C & 134 & ViTamin-L2-datacomp1b & purple / purple / — / purple & C \\
135 & ViTamin-S-LTT-datacomp1b & — / — / achromatic / achromatic & A & 136 & ViTamin-S-datacomp1b & — / — / — / — & B \\
137 & ViTamin-XL-256-datacomp1b & purple / — / green / purple & C & 138 & coca\_ViT-B-32-laion2b\_s13b\_b90k & blue / blue / blue / blue & C \\
139 & coca\_ViT-B-32-mscoco\_finetuned\_laion2b\_s13b\_b90k & blue / — / blue / blue & C & 140 & coca\_ViT-L-14-laion2b\_s13b\_b90k & achromatic / — / blue / blue & C \\
141 & coca\_ViT-L-14-mscoco\_finetuned\_laion2b\_s13b\_b90k & blue / — / achromatic / — & B & 142 & convnext\_base-laion400m\_s13b\_b51k & pink / blue / — / — & B \\
143 & convnext\_base\_w-laion2b\_s13b\_b82k & achromatic / blue / achromatic / blue & A & 144 & convnext\_base\_w-laion2b\_s13b\_b82k\_augreg & — / — / blue / — & B \\
145 & convnext\_base\_w-laion\_aesthetic\_s13b\_b82k & achromatic / blue / — / achromatic & A & 146 & convnext\_base\_w\_320-laion\_aesthetic\_s13b\_b82k & achromatic / achromatic / — / — & A \\
147 & convnext\_base\_w\_320-laion\_aesthetic\_s13b\_b82k\_augreg & red / — / — / — & B & 148 & convnext\_large\_d-laion2b\_s26b\_b102k\_augreg & — / red / — / — & B \\
149 & convnext\_large\_d\_320-laion2b\_s29b\_b131k\_ft & — / red / blue / — & B & 150 & convnext\_large\_d\_320-laion2b\_s29b\_b131k\_ft\_soup & red / red / blue / blue & C \\
151 & convnext\_xxlarge-laion2b\_s34b\_b82k\_augreg & — / — / blue / blue & C & 152 & convnext\_xxlarge-laion2b\_s34b\_b82k\_augreg\_rewind & green / — / blue / blue & C \\
153 & convnext\_xxlarge-laion2b\_s34b\_b82k\_augreg\_soup & — / — / blue / blue & C & 154 & nllb-clip-base-siglip-mrl & — / — / green / — & B \\
155 & nllb-clip-base-siglip-v1 & — / — / — / — & B & 156 & nllb-clip-base-v1 & — / — / — / — & B \\
157 & nllb-clip-large-siglip-mrl & — / blue / yellow / — & B & 158 & nllb-clip-large-siglip-v1 & — / — / — / achromatic & B \\
159 & nllb-clip-large-v1 & — / orange / — / — & B & 160 & roberta-ViT-B-32-laion2b\_s12b\_b32k & yellow / — / — / — & B \\
161 & xlm-roberta-base-ViT-B-32-laion5b\_s13b\_b90k & achromatic / — / achromatic / — & A & 162 & xlm-roberta-large-ViT-H-14-frozen\_laion5b\_s13b\_b90k & achromatic / achromatic / — / achromatic & A \\
\bottomrule
\end{tabular}
\end{adjustbox}
\end{table*}

\section{Robustness of the Human Baseline to Uncalibrated Displays}
\label{app:Adaptation}

A potential limitation of crowdsourcing psychophysical data is the lack of strict hardware control, as participants performed the experiment on various uncalibrated personal displays. To validate the robustness of our human baseline, we hypothesize that the physical chromatic shifts introduced by different monitors are analogous to natural changes in environmental lighting—a phenomenon the human visual system seamlessly resolves through chromatic adaptation. To test this hypothesis, we extracted a representative sample of 16 colors from the Hues and Cues board. We then generated two experimental scenarios:
\begin{itemize}
    \item Display Variance: The chromatic coordinates of the 16 colors were measured across three different uncalibrated monitors.
    \item Illuminant Variance: The same 16 colors were mathematically simulated under three illuminants with tristimulus values that match the tristimulus values of the achromatic sample and average color shown in the displays.
\end{itemize}

To simulate human color constancy and asses the dificulty of the color compensation problem, (1) we applied a standard Von Kries chromatic adaptation transform to the data from both scenarios, and (2) We evaluated the statistical equivalence of the shifts in the color distributions using Analysis of Variance (ANOVA).

The statistical analysis yielded two key findings. First, an ANOVA test comparing the unadapted chromatic coordinates of the displays versus the unadapted coordinates of the illuminants revealed no significant difference between them (the equal distribution null-hypothesis cannot be rejected $p = 0.1102$). This confirms that the chromatic distortion caused by hardware variance is statistically equivalent to a standard change in lighting. Second, an ANOVA test comparing the coordinates after applying the Von Kries adaptation in both scenarios also showed no significant difference ($p = 0.3333$).

\begin{figure*}[h]
    \centering
    \includegraphics[width=1\linewidth]{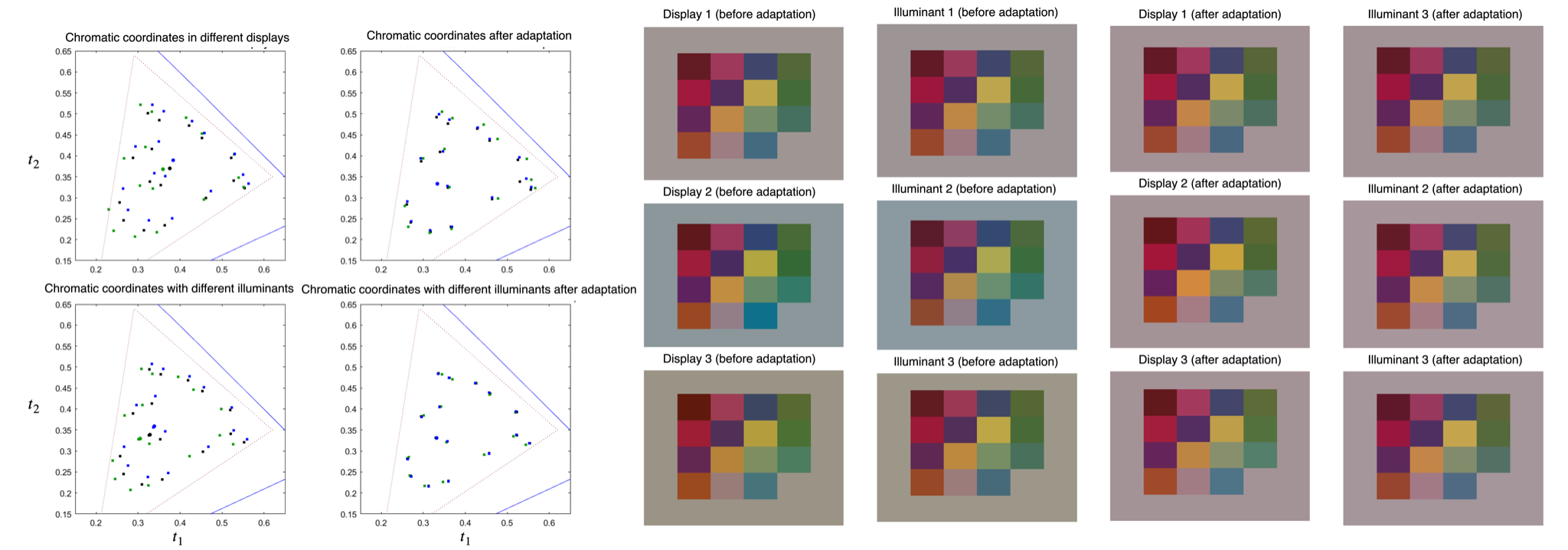}
    
    \caption{Validation of display robustness via chromatic adaptation. The left panels display the CIE $xy$ chromatic coordinates of a 16-color sample under uncalibrated displays (top row) and different illuminants (bottom row), both before and after applying a Von Kries chromatic adaptation transform. The right panels visualize the corresponding rendered color patches for these four scenarios. ANOVA testing confirms that hardware variance is statistically equivalent to illuminant variance ($p = 0.0914$) and that chromatic adaptation normalizes both scenarios equally ($p = 0.3214$), ensuring that human observers perceive consistent color categories regardless of the specific device used.}
    \label{fig:Adaptaciones}
\end{figure*}

These results, visualized in Figure~\ref{fig:Adaptaciones}, empirically demonstrate that the human visual system's natural chromatic adaptation effectively normalizes the physical deviations introduced by personal screens. Consequently, the semantic categorization of the board colors remains robust and consistent across different devices, validating our remote data acquisition methodology.

\section{Complete Word-Level Alignment and Semantic Categorization}
\label{app: Word_color}

This appendix provides the exhaustive word-level breakdown of the experiments discussed in Section~\ref{sec:results}. Table~\ref{tab:all_categories} details the aggregated predictions of the evaluated Contrastive Vision-Language Models compared against the human consensus for the entire 100-word vocabulary. To facilitate a structured analysis of semantic grounding, the data is segregated into the seven predefined categories: Food, Animal, Plant, Mineral, Environment, Subjective, and Pop Culture.

For each target concept, we report the empirical Human Consistency. The predictions for both the human distribution and the model average are represented by their discrete physical coordinate on the gamified board (e.g., A13, P10). Furthermore, to aid qualitative interpretation, these coordinates are mapped to their corresponding macro-color categorization, as established in the tessellation detailed in Section~\ref{sec: color_naming}.  

This granular presentation serves a dual purpose. First, it extensively documents the models' capacity to align with idealized human memory colors for concrete physical referents (e.g., yielding robust consensus in the Food and Plant categories). Second, it exposes the severity of the misalignments in abstract domains, explicitly illustrating the systematic uncertainty collapse into the default blue coordinate (I18) previously addressed in Section~\ref{sec:results_global}.

\begin{table*}[h]
\centering
\caption{Comprehensive breakdown of model predictions versus human consensus across all semantic categories. The empirical baseline is reported as Human Consistency, reflecting the intra-class agreement among human observers.}
\label{tab:all_categories}
\resizebox{\textwidth}{!}{
\begin{tabular}{llcllll | llcllll}
\toprule
\multirow{2}{*}{\textbf{Category}} & \multirow{2}{*}{\textbf{Word}} & \multirow{2}{*}{\textbf{Consistency}} & \multicolumn{2}{c}{\textbf{Humans}} & \multicolumn{2}{c|}{\textbf{Models}} & \multirow{2}{*}{\textbf{Category}} & \multirow{2}{*}{\textbf{Word}} & \multirow{2}{*}{\textbf{Consistency}} & \multicolumn{2}{c}{\textbf{Humans}} & \multicolumn{2}{c}{\textbf{Models}} \\
\cmidrule{4-7} \cmidrule{11-14}
& & & \textbf{Coord.} & \textbf{Mem. Color} & \textbf{Coord.} & \textbf{Mem. Color} & & & & \textbf{Coord.} & \textbf{Mem. Color} & \textbf{Coord.} & \textbf{Mem. Color} \\
\midrule
\textbf{Food} & APPLE & 0.89 & \cellcolor[RGB]{194,54,46} A13 & Red & \cellcolor[RGB]{191,193,194} I18 & Blue & \textbf{Environment} & BLOOD & 0.89 & \cellcolor[RGB]{165,53,45} A8 & Red & \cellcolor[RGB]{130,45,39} A6 & Brown \\
 & ASPARAGUS & 0.83 & \cellcolor[RGB]{65,118,66} P10 & Green & \cellcolor[RGB]{112,140,60} P1 & Green &  & FIRE & 0.88 & \cellcolor[RGB]{206,66,49} B10 & Red & \cellcolor[RGB]{205,72,45} C8 & Red \\
 & BANANA & 0.90 & \cellcolor[RGB]{238,198,69} I4 & Yellow & \cellcolor[RGB]{238,220,101} J7 & Yellow &  & GRASS & 0.90 & \cellcolor[RGB]{106,159,69} N10 & Green & \cellcolor[RGB]{102,139,65} P4 & Green \\
 & BEET & 0.90 & \cellcolor[RGB]{175,49,110} A20 & Pink & \cellcolor[RGB]{175,49,110} A20 & Pink &  & HURRICANE & 0.82 & \cellcolor[RGB]{73,99,146} L29 & Blue & \cellcolor[RGB]{191,193,194} I18 & Blue \\
 & COCONUT & 0.89 & \cellcolor[RGB]{90,47,30} A2 & Brown & \cellcolor[RGB]{82,48,27} A1 & Brown &  & MOSS & 0.83 & \cellcolor[RGB]{63,120,66} P11 & Green & \cellcolor[RGB]{112,140,60} P1 & Green \\
 & CUCUMBER & 0.87 & \cellcolor[RGB]{65,118,66} P10 & Green & \cellcolor[RGB]{152,186,114} L12 & Green &  & MUD & 0.84 & \cellcolor[RGB]{82,48,27} A1 & Brown & \cellcolor[RGB]{82,48,27} A1 & Brown \\
 & EGG & 0.90 & \cellcolor[RGB]{213,114,50} E6 & Orange & \cellcolor[RGB]{238,220,101} J7 & Yellow &  & POND & 0.94 & \cellcolor[RGB]{109,175,158} M22 & Blue & \cellcolor[RGB]{112,140,60} P1 & Green \\
 & EGGPLANT & 0.87 & \cellcolor[RGB]{83,52,121} E28 & Purple & \cellcolor[RGB]{111,59,115} B27 & Purple &  & SAND & 0.90 & \cellcolor[RGB]{232,172,88} G5 & Orange & \cellcolor[RGB]{207,190,146} I13 & Green \\
 & GRAPE & 0.95 & \cellcolor[RGB]{155,97,139} E20 & Purple & \cellcolor[RGB]{131,58,113} A25 & Purple &  & SEA & 0.94 & \cellcolor[RGB]{75,128,163} N29 & Blue & \cellcolor[RGB]{93,172,174} M24 & Blue \\
 & KIWI & 0.93 & \cellcolor[RGB]{82,48,27} A1 & Brown & \cellcolor[RGB]{161,186,60} N4 & Green &  & SKIN & 0.86 & \cellcolor[RGB]{82,48,27} A1 & Brown & \cellcolor[RGB]{227,160,127} H12 & Orange \\
 & LEMON & 0.89 & \cellcolor[RGB]{228,217,53} L2 & Yellow & \cellcolor[RGB]{235,222,79} K4 & Yellow &  & SKY & 0.90 & \cellcolor[RGB]{130,186,197} K21 & Blue & \cellcolor[RGB]{96,164,187} L24 & Blue \\
 & LIME & 0.89 & \cellcolor[RGB]{194,204,68} M5 & Green & \cellcolor[RGB]{180,193,51} N1 & Green &  & SUNRISE & 0.85 & \cellcolor[RGB]{233,174,50} G2 & Yellow & \cellcolor[RGB]{235,159,109} H10 & Orange \\
 & PEACH & 0.86 & \cellcolor[RGB]{223,136,44} E3 & Orange & \cellcolor[RGB]{225,146,123} G11 & Pink &  & SUNSET & 0.87 & \cellcolor[RGB]{205,95,49} D8 & Orange & \cellcolor[RGB]{235,159,109} H10 & Orange \\
 & PUMPKIN & 0.86 & \cellcolor[RGB]{218,131,42} E4 & Orange & \cellcolor[RGB]{223,136,44} E3 & Orange &  & THUNDERBOLT & 0.82 & \cellcolor[RGB]{234,219,67} K2 & Yellow & \cellcolor[RGB]{191,193,194} I18 & Blue \\
\cmidrule{8-14}
 & TOMATO & 0.80 & \cellcolor[RGB]{202,63,44} B9 & Red & \cellcolor[RGB]{213,78,66} C10 & Red & \textbf{Subjective} & AUTUMN & 0.92 & \cellcolor[RGB]{128,70,41} B2 & Brown & \cellcolor[RGB]{191,112,45} D3 & Brown \\
 & ZUCCHINI & 0.86 & \cellcolor[RGB]{65,118,66} P10 & Green & \cellcolor[RGB]{147,170,59} O1 & Green &  & DANGER & 0.81 & \cellcolor[RGB]{194,54,46} A13 & Red & \cellcolor[RGB]{191,193,194} I18 & Blue \\
\cmidrule{1-7}
\textbf{Animal} & AXOLOTL & 0.89 & \cellcolor[RGB]{222,144,155} G15 & Pink & \cellcolor[RGB]{191,193,194} I18 & Blue &  & DISGUST & 0.82 & \cellcolor[RGB]{65,118,66} P10 & Green & \cellcolor[RGB]{191,193,194} I18 & Blue \\
 & CANARY & 0.81 & \cellcolor[RGB]{239,213,61} J3 & Yellow & \cellcolor[RGB]{229,216,45} L1 & Yellow &  & ECOLOGY & 0.87 & \cellcolor[RGB]{77,150,66} O14 & Green & \cellcolor[RGB]{126,173,68} N8 & Green \\
 & CRICKET & 0.88 & \cellcolor[RGB]{65,118,66} P10 & Green & \cellcolor[RGB]{126,173,68} N8 & Green &  & HAPINESS & 0.86 & \cellcolor[RGB]{224,203,44} J1 & Yellow & \cellcolor[RGB]{191,193,194} I18 & Blue \\
 & FLAMINGO & 0.82 & \cellcolor[RGB]{218,127,144} F15 & Pink & \cellcolor[RGB]{217,110,120} E14 & Pink &  & HOPE & 0.87 & \cellcolor[RGB]{91,158,86} N14 & Green & \cellcolor[RGB]{191,193,194} I18 & Blue \\
 & FOX & 0.89 & \cellcolor[RGB]{211,99,50} D7 & Orange & \cellcolor[RGB]{211,99,50} D7 & Orange &  & LONELINESS & 0.78 & \cellcolor[RGB]{51,47,105} G30 & Blue & \cellcolor[RGB]{191,193,194} I18 & Blue \\
 & FROG & 0.87 & \cellcolor[RGB]{63,120,66} P11 & Green & \cellcolor[RGB]{130,175,68} N7 & Green &  & PEACE & 0.81 & \cellcolor[RGB]{191,193,194} I18 & Blue & \cellcolor[RGB]{191,193,194} I18 & Blue \\
 & GRIZZLY & 0.85 & \cellcolor[RGB]{82,48,27} A1 & Brown & \cellcolor[RGB]{82,48,27} A1 & Brown &  & RAGE & 0.85 & \cellcolor[RGB]{194,54,46} A13 & Red & \cellcolor[RGB]{191,193,194} I18 & Blue \\
 & LADYBUG & 0.88 & \cellcolor[RGB]{194,54,46} A13 & Red & \cellcolor[RGB]{205,59,66} A15 & Red &  & RELAX & 0.84 & \cellcolor[RGB]{191,193,194} I18 & Blue & \cellcolor[RGB]{191,193,194} I18 & Blue \\
 & PIG & 0.89 & \cellcolor[RGB]{218,127,144} F15 & Pink & \cellcolor[RGB]{218,165,158} H15 & Pink &  & SADNESS & 0.85 & \cellcolor[RGB]{69,114,155} N30 & Blue & \cellcolor[RGB]{191,193,194} I18 & Blue \\
 & SALMON & 0.91 & \cellcolor[RGB]{209,93,76} D10 & Red & \cellcolor[RGB]{223,134,112} F11 & Pink &  & SHAME & 0.89 & \cellcolor[RGB]{214,104,131} E15 & Pink & \cellcolor[RGB]{191,193,194} I18 & Blue \\
 & TIGER & 0.86 & \cellcolor[RGB]{213,114,50} E6 & Orange & \cellcolor[RGB]{223,136,44} E3 & Orange &  & SPRING & 0.88 & \cellcolor[RGB]{214,104,131} E15 & Pink & \cellcolor[RGB]{104,176,110} N20 & Green \\
 & TOAD & 0.80 & \cellcolor[RGB]{65,118,66} P10 & Green & \cellcolor[RGB]{191,193,194} I18 & Blue &  & SUMMER & 0.82 & \cellcolor[RGB]{238,217,81} J5 & Yellow & \cellcolor[RGB]{101,176,189} L23 & Blue \\
 & WHALE & 0.87 & \cellcolor[RGB]{85,118,157} K26 & Blue & \cellcolor[RGB]{191,193,194} I18 & Blue &  & TOXIC & 0.87 & \cellcolor[RGB]{229,216,45} L1 & Yellow & \cellcolor[RGB]{180,193,51} N1 & Green \\
\cmidrule{1-7}
\textbf{Plant} & BASIL & 0.88 & \cellcolor[RGB]{78,140,69} O12 & Green & \cellcolor[RGB]{145,178,120} L13 & Green &  & WINTER & 0.90 & \cellcolor[RGB]{96,164,187} L24 & Blue & \cellcolor[RGB]{191,193,194} I18 & Blue \\
\cmidrule{8-14}
 & CARNATION & 0.95 & \cellcolor[RGB]{204,68,122} C17 & Pink & \cellcolor[RGB]{222,144,155} G15 & Pink & \textbf{Pop Culture} & BARBIE & 0.84 & \cellcolor[RGB]{209,65,110} C16 & Pink & \cellcolor[RGB]{204,68,122} C17 & Pink \\
 & CORAL & 0.92 & \cellcolor[RGB]{210,75,50} C9 & Red & \cellcolor[RGB]{223,134,112} F11 & Pink &  & BERT (AND ERNIE) & 0.86 & \cellcolor[RGB]{217,125,46} E5 & Orange & \cellcolor[RGB]{191,193,194} I18 & Blue \\
 & LAVENDER & 0.90 & \cellcolor[RGB]{145,64,120} B23 & Purple & \cellcolor[RGB]{142,131,158} H21 & Purple &  & BRUSH & 0.85 & \cellcolor[RGB]{214,104,131} E15 & Pink & \cellcolor[RGB]{191,193,194} I18 & Blue \\
 & MALLOW & 0.88 & \cellcolor[RGB]{204,68,122} C17 & Pink & \cellcolor[RGB]{192,166,175} H17 & Purple &  & COOKIE MONSTER & 0.90 & \cellcolor[RGB]{48,65,124} I30 & Blue & \cellcolor[RGB]{77,141,173} N28 & Blue \\
 & MINT & 0.89 & \cellcolor[RGB]{65,118,66} P10 & Green & \cellcolor[RGB]{140,190,165} L19 & Green &  & ELMO & 0.81 & \cellcolor[RGB]{205,61,45} A10 & Red & \cellcolor[RGB]{205,61,45} A10 & Red \\
 & POPPY & 0.89 & \cellcolor[RGB]{192,58,44} A9 & Red & \cellcolor[RGB]{205,61,45} A10 & Red &  & ERNIE (AND BERT) & 0.83 & \cellcolor[RGB]{237,216,75} J4 & Yellow & \cellcolor[RGB]{191,193,194} I18 & Blue \\
 & SUNFLOWER & 0.90 & \cellcolor[RGB]{238,198,69} I4 & Yellow & \cellcolor[RGB]{239,203,76} I5 & Yellow &  & FEMINISM & 0.85 & \cellcolor[RGB]{97,59,117} A30 & Purple & \cellcolor[RGB]{199,54,115} B18 & Pink \\
 & WHEAT & 0.91 & \cellcolor[RGB]{230,162,42} F2 & Orange & \cellcolor[RGB]{207,190,146} I13 & Green &  & FLASH & 0.81 & \cellcolor[RGB]{209,65,59} B12 & Red & \cellcolor[RGB]{203,57,56} A14 & Red \\
 & WOOD & 0.85 & \cellcolor[RGB]{115,71,37} B1 & Brown & \cellcolor[RGB]{143,92,43} C1 & Brown &  & GRINCH & 0.86 & \cellcolor[RGB]{130,175,68} N7 & Green & \cellcolor[RGB]{154,184,62} N5 & Green \\
\cmidrule{1-7}
\textbf{Mineral} & AMBER & 0.89 & \cellcolor[RGB]{218,131,42} E4 & Orange & \cellcolor[RGB]{217,140,41} E2 & Orange &  & HULK & 0.75 & \cellcolor[RGB]{65,118,66} P10 & Green & \cellcolor[RGB]{126,173,68} N8 & Green \\
 & AMETHYST & 0.85 & \cellcolor[RGB]{83,52,121} E28 & Purple & \cellcolor[RGB]{97,59,117} A30 & Purple &  & PETER PAN & 0.88 & \cellcolor[RGB]{106,167,105} N17 & Green & \cellcolor[RGB]{70,145,67} P15 & Green \\
 & BRICK & 0.83 & \cellcolor[RGB]{196,61,46} B8 & Red & \cellcolor[RGB]{150,70,45} B4 & Brown &  & PIKACHU & 0.83 & \cellcolor[RGB]{239,213,61} J3 & Yellow & \cellcolor[RGB]{239,213,61} J3 & Yellow \\
 & EMERALD & 0.86 & \cellcolor[RGB]{116,167,70} N9 & Green & \cellcolor[RGB]{63,120,66} P11 & Green &  & SHRECK & 0.85 & \cellcolor[RGB]{119,172,112} M15 & Green & \cellcolor[RGB]{154,184,62} N5 & Green \\
 & GOLD & 0.94 & \cellcolor[RGB]{230,186,38} H1 & Yellow & \cellcolor[RGB]{221,168,33} G1 & Yellow &  & SIMPSON & 0.85 & \cellcolor[RGB]{235,211,51} J2 & Yellow & \cellcolor[RGB]{229,216,45} L1 & Yellow \\
 & JADE & 0.91 & \cellcolor[RGB]{77,150,66} O14 & Green & \cellcolor[RGB]{87,159,119} P24 & Green &  & SMURF & 0.88 & \cellcolor[RGB]{101,176,189} L23 & Blue & \cellcolor[RGB]{77,141,173} N28 & Blue \\
 & LAPIS LAZULI & 0.92 & \cellcolor[RGB]{48,65,124} I30 & Blue & \cellcolor[RGB]{49,56,116} H30 & Blue &  & SONIC & 0.84 & \cellcolor[RGB]{48,65,124} I30 & Blue & \cellcolor[RGB]{48,65,124} I30 & Blue \\
 & PEARL & 0.83 & \cellcolor[RGB]{191,193,194} I18 & Blue & \cellcolor[RGB]{194,198,180} I16 & Green &  & (LILO AND) STITCH & 0.90 & \cellcolor[RGB]{81,147,175} M26 & Blue & \cellcolor[RGB]{191,193,194} I18 & Blue \\
 & RUBY & 0.85 & \cellcolor[RGB]{196,61,46} B8 & Red & \cellcolor[RGB]{197,53,86} A17 & Pink &  & SUBMARINE & 0.88 & \cellcolor[RGB]{235,211,51} J2 & Yellow & \cellcolor[RGB]{191,193,194} I18 & Blue \\
 & SAPPHIRE & 0.86 & \cellcolor[RGB]{48,65,124} I30 & Blue & \cellcolor[RGB]{48,65,124} I30 & Blue &  & TRACTOR & 0.78 & \cellcolor[RGB]{239,213,61} J3 & Yellow & \cellcolor[RGB]{191,193,194} I18 & Blue \\
 & TILE & 0.86 & \cellcolor[RGB]{143,48,41} A7 & Brown & \cellcolor[RGB]{192,196,189} I17 & Blue &  & WITCHCRAFT & 0.81 & \cellcolor[RGB]{82,48,27} A1 & Brown & \cellcolor[RGB]{191,193,194} I18 & Blue \\
\bottomrule
\end{tabular}
}
\end{table*}

\section{Comprehensive Nonsense Baseline Responses Across All Evaluated Models}
\label{app: Colapso}

This appendix provides the complete visual breakdown of the "Nonsense Baseline" experiment discussed in Section~\ref{sec:results_collapse}. The purpose of this exhaustive visualization is to document the intrinsic chromatic priors and uncertainty behaviors of all 162 Contrastive Vision-Language Models evaluated in our study.

\begin{figure*}[!h]
	\centering
	\includegraphics[width=1\linewidth]{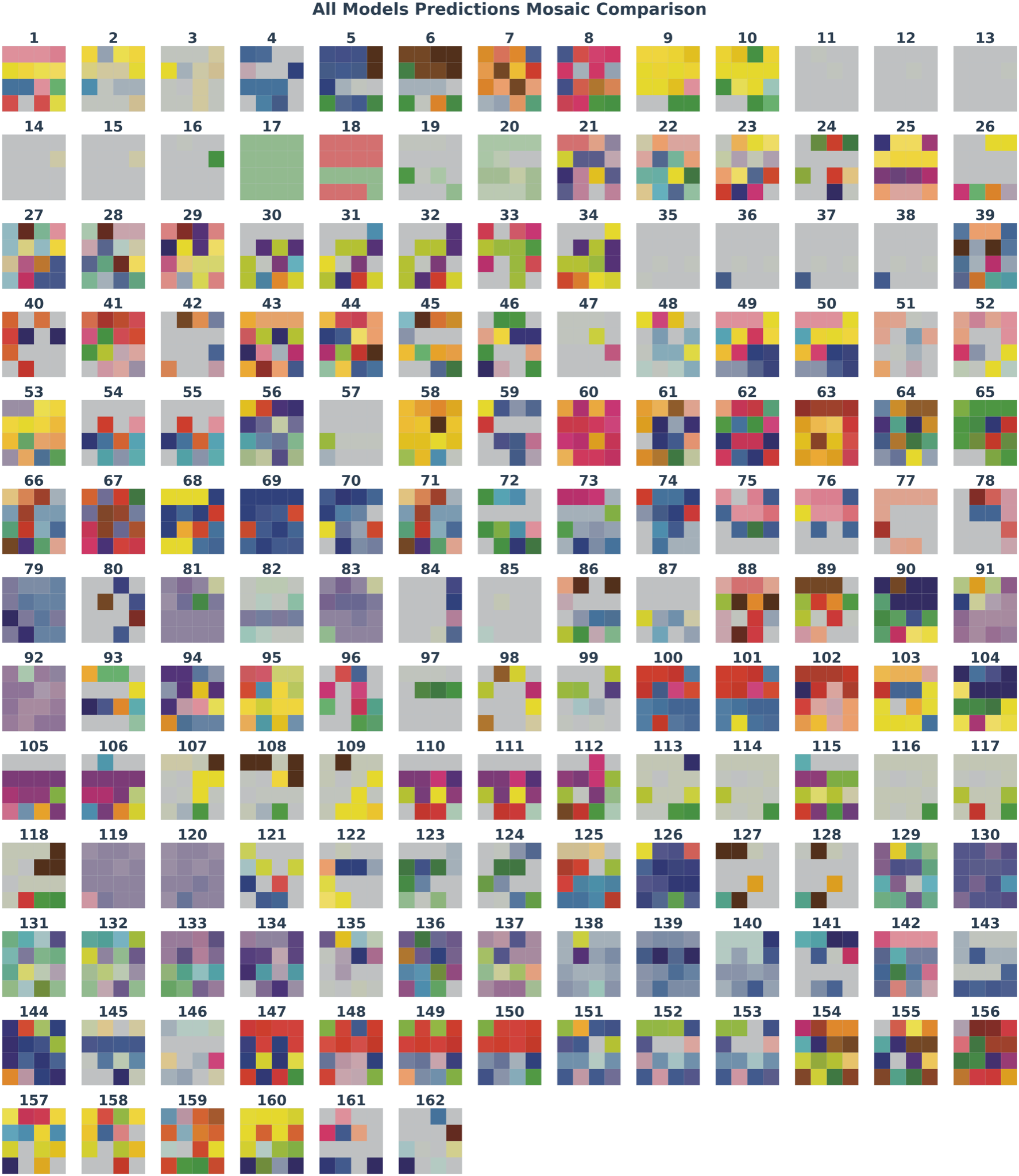}
	\caption{Visualization of the "Nonsense Baseline" responses across all evaluated models. Each block represents the Top-1 color prediction for a model when subjected to semantically void inputs, as defined in Figure \ref{fig:TipoColapso} (Numbers, Stop Words, Pseudowords, and Random Strings). The mosaic illustrates the models' internal priors: while some architectures exhibit high-entropy (random) behavior, a significant portion displays a systematic "Blue Bias," collapsing toward specific chromatic regions in the absence of valid semantic referents. This confirms that the models' error patterns in valid vocabulary are often reflections of these intrinsic priors rather than learned semantic misassociations.}
	\label{fig:GlobalColapso}
\end{figure*}

Figure~\ref{fig:GlobalColapso} illustrates the Top-1 chromatic predictions for each model when subjected to 16 semantically void inputs. Each individual $4\times4$ mosaic corresponds to a single CVLM. The spatial arrangement within each mosaic maps directly to the stimuli legend defined previously (comprising Numbers, Stop Words, Pseudowords, and Random Strings).

By examining this global compilation, readers can observe the full spectrum of the three behavioral modes identified in our analysis, as can be seen in Figure~\ref{fig:TipoColapso}. 

We provide this comprehensive reference so that researchers and practitioners can identify the specific biases of individual architectures. Recognizing these model-specific priors is a critical step in visual alignment evaluation, as it allows researchers to flag potential false positives—instances where a model's apparent semantic success is merely an artifact of its underlying chromatic bias.

\printcredits

\bibliographystyle{elsarticle-harv}

\bibliography{cas-refs}







\end{document}